\documentclass[]{bytedance_seed}

\usepackage[toc,page,header]{appendix}
\usepackage{amsmath,amssymb,amsthm}
\usepackage{microtype}
\usepackage{capt-of}
\usepackage{adjustbox}
\usepackage{colortbl}

\input{common}
\newcommand{\dd}{\mathrm{d}}
\newcommand{\E}{\mathbb{E}}
\newcommand{\R}{\mathbb{R}}
\newcommand{\norm}[1]{\lVert #1\rVert}

\newcommand{\elr}{\eta^{\mathrm{eff}}}
\newcommand{\Loss}{\mathcal{L}}
\newcommand{\hist}{\mathcal{F}}

\title{\mytitle}
\hypersetup{
    pdftitle={\mytitleplain},
    pdfauthor={Qifan Yu, Xinyu Ma, Xingyan Bin, Di He}
}

\author[1,2,*]{Qifan Yu}
\author[1]{Xinyu Ma}
\author[1,\dag]{Xingyan Bin}
\author[2,\dag]{Di He}

\affiliation[1]{ByteDance Seed}
\affiliation[2]{Peking University}

\contribution[*]{Work done at ByteDance Seed}
\contribution[\dag]{Corresponding authors}

\abstract{
    Weight-normalized (WN) training is emerging as a compelling alternative to
conventional weight decay (WD), with recent studies reporting substantial
pretraining loss gains.
We find that the effective learning rate (ELR) trajectory accounts
for most of the loss dynamics difference between WN and WD.
Matching blockwise ELR trajectories step by step closely aligns their loss curves.
\emph{Schedule design should therefore target ELR rather than the nominal
learning rate (LR).}
Building on this principle, we model directional-alignment dynamics under WN and derive $1/\sqrt{t}$ decay as the optimal schedule.
For finite-horizon pretraining, we propose \textbf{\OurMethod} (\OurMethodDef) to match prescribed peak and terminal LRs.
Experiments on 128\,B- and 500\,B-token Mixture-of-Experts pretraining under WN show that \OurMethod outperforms cosine and alternative schedules in loss, downstream performance, and hyperparameter robustness, reducing terminal training loss by 0.01 relative to cosine.
We further identify late-stage scheduling as the primary source of this gain through controlled stage interventions, and demonstrate that a shared late schedule progressively attenuates alignment differences left by earlier LRs.
Finally, transferring \OurMethod's ELR trajectory to WD reproduces the same improved loss dynamics, demonstrating that these gains come from ELR scheduling itself and extend beyond fixed-norm training.

}

\date{\today}
\correspondence{Qifan Yu at \email{qifanyu@stu.pku.edu.cn}, Xinyu Ma at \email{maxinyu.flame@bytedance.com}, \\
\hphantom{{\bffont\bfseries\sffamily\footnotesize Correspondence:}} Xingyan Bin at \email{binxingyan@bytedance.com}, Di He at \email{di\_he@pku.edu.cn}}

\begin{document}
\maketitle

\section{Introduction}
\label{sec:intro}
\begin{figure}[htbp]
    \centering
    \includegraphics[width=.95\textwidth]{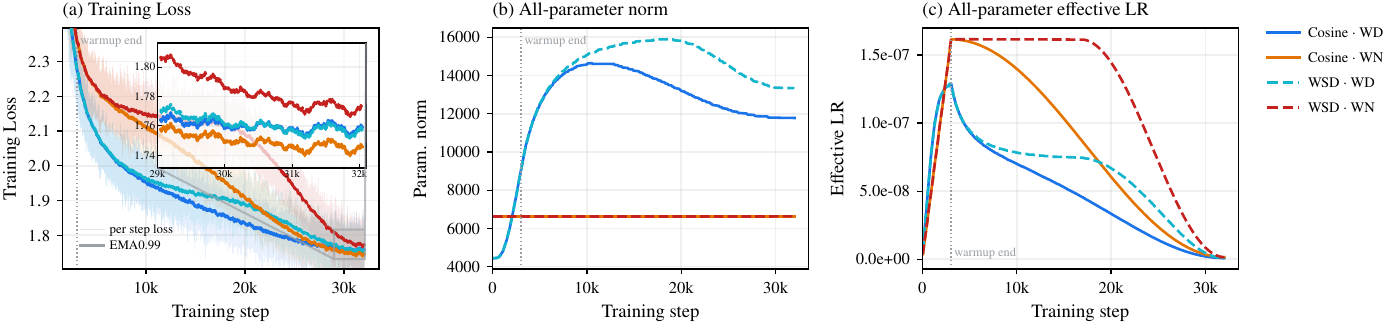}
    \vspace{ - 2pt}
    \caption{\textbf{WN overtakes WD under cosine but remains behind under WSD.}
    Panel~\textbf{(a)} shows training loss, panel~\textbf{(b)} the all-parameter norm, and panel~\textbf{(c)} the aggregate effective LR, defined as the nominal LR divided by the all-parameter norm.
    }
    \label{fig:elr_cosine_wsd}
\end{figure}

In large language model (LLM) pretraining, weight-normalized (WN) training~\citep{salimans2016wn,wen2026hyperball} has emerged as a promising alternative to conventional weight decay (WD)~\citep{loshchilov2019adamw}.
It fixes parameter norms after each update, with recent work reporting lower pretraining loss, improved token efficiency, greater stability, and better hyperparameter transfer~\citep{wen2026hyperball,loshchilov2026ngpt,an2026demystifyingmanifoldconstraintsllm}.
Yet WN and WD exhibit markedly different training dynamics under the same nominal learning rate (LR) schedule.
In Figure~\ref{fig:elr_cosine_wsd}(a), under cosine decay~\citep{loshchilov2017sgdr}, WN initially trails WD but eventually overtakes it.
However, under warmup-stable-decay (WSD)~\citep{hu2024minicpm}, WN remains behind WD throughout training and ends with substantially higher loss.
Explaining why their loss ranking differs across schedules is essential to understanding when and why WN outperforms WD.

To explain this contrast, we examine the \emph{effective learning rate} (ELR)~\citep{vanlaarhoven2017l2,dangelo2024weightdecay}, defined for each parameter block as \(\eta^{\mathrm{eff}}=\eta/r\), which reflects the effective update scale relative to the block's norm \(r\) under nominal LR \(\eta\).
A shared nominal LR schedule can induce distinct ELR trajectories when norms evolve differently.
During the stable phase of WSD, the nominal LR is constant, so ELR changes reflect norm evolution alone (Figure~\ref{fig:elr_cosine_wsd}(b,c)):
WN fixes \(r\) and holds ELR at peak, leading to a slow loss decrease.
In contrast, WD's growing norm lowers ELR, coinciding with its faster loss drop.
Under cosine decay, however,
WD's growing norm drives ELR down rapidly early on and slows late progress, whereas WN sustains higher early ELR and continues annealing later to overtake WD.
Loss dynamics therefore track the \emph{full ELR trajectory} across training rather than instantaneous ELR magnitude.

Inspired by these observations, we further validate the role of the ELR trajectory through controlled, bidirectional blockwise interventions between WN and WD (Section~\ref{sec:elr_principle_alignment}).
Specifically, once we match their blockwise ELR trajectories step by step, their loss dynamics collapse onto nearly the same curve even when the runs follow substantially different parameter norms and nominal LRs (Figure~\ref{fig:elr_dynamics_collapse}).
Concurrent studies~\citep{liu2026elr,xiao2026hyperball,zhou-zhou-gu-2026-elr} also report similar results, corroborating our findings.

Taken together, these findings establish the induced ELR trajectory, the per-step \(\eta/r\) sequence jointly set by nominal LR and parameter norms, as a key determinant of loss dynamics.
\textbf{Principled LR schedule design should therefore target \emph{ELR} rather than the nominal LR
alone.}
This core idea naturally inspires an essential research question:
\begin{center}
\emph{What should a better ELR schedule look like?}
\end{center}
To address this question, we study ELR scheduling under WN, where fixing each block's norm at \(R\) decouples \(\eta^{\mathrm{eff}}=\eta/R\) from norm evolution and makes ELR directly proportional to nominal LR.
Starting from the WN update rule, we model directional-alignment dynamics toward the optimal parameter direction on the sphere (Section~\ref{sec:wn_schedule}).
When no peak or terminal LR is prescribed, maximizing expected alignment yields $1/\sqrt{t}$ decay, under conditions that we verify empirically in MoE pretraining.
To adapt this schedule to practical pretraining, which decays to a prescribed terminal LR over a finite horizon, we introduce \textbf{\OurMethod} (\OurMethodDef), an affine transformation that matches prescribed peak and terminal LRs while preserving the relative decay rates of $1/\sqrt{t}$ across steps.

Empirically, in 128\,B- and 500\,B-token Mixture-of-Experts (MoE)~\citep{shazeer2017moe} pretraining, we evaluate \OurMethod under WN against cosine and alternative LR schedule families (Section~\ref{sec:new_evaluation}).
\OurMethod reduces terminal training loss by approximately 0.01 relative to cosine, achieving the lowest loss, the best downstream performance, and greatest robustness to hyperparameter choices.
Controlled stage swaps and shared-tail studies further attribute most of \OurMethod's gain to late-stage scheduling (Section~\ref{sec:new_late_stage}): sharing the late-stage \OurMethod schedule removes most of the final loss gap across distinct early schedules.
Our alignment theory explains this result by showing that a shared late schedule progressively attenuates alignment differences left by earlier LRs.
Finally, transferring \OurMethod's ELR trajectory to WD reproduces the same improved loss dynamics, demonstrating that these gains come from ELR scheduling itself and can be extended beyond fixed-norm training.

\newpage
In summary, our contributions are:
\begin{itemize}
    \item We show via bidirectional blockwise ELR alignment that the \textbf{induced ELR trajectory accounts for most of the observed loss dynamics difference} between WN and WD, motivating ELR rather than nominal LR as the scheduling target.

    \item Modeling directional-alignment dynamics under WN yields $1/\sqrt{t}$ decay, from which we propose \textbf{\OurMethod, an affine $1/\sqrt{t}$ schedule} for finite-horizon pretraining.

    \item Across 128\,B- and 500\,B-token MoE pretraining under WN, \textbf{\OurMethod outperforms cosine and alternative schedules} in loss, downstream accuracy, and hyperparameter robustness, with gains driven primarily by late-stage scheduling and transferable to WD.
\end{itemize}

\section{Effective Learning Rate as the Scheduling Principle}
\label{sec:elr_principle}


\subsection{Effective learning rate under WN and WD}
\label{sec:elr_principle_updates}

Let \(\bm\theta_k\) denote a 
parameter block at optimizer step \(k\), \(\eta_k\geq 0\) its nominal learning rate, and \(\bm u_k\) the update before LR scaling.
With decay coefficient \(\lambda\geq0\) and prescribed norm \(R>0\), decoupled WD~\citep{loshchilov2019adamw} and fixed-norm WN~\citep{wen2026hyperball} update the block as
\begin{align}
    \text{WD:}\quad
    \bm\theta_{k+1}
    &= (1-\lambda\eta_k)\bm\theta_k+\eta_k\bm u_k,
    \label{eq:principle_wd_update}\\
    \text{WN:}\quad
    \bm\theta_{k+1}
    &= R\frac{\bm\theta_k+\eta_k\bm u_k}
    {\norm{\bm\theta_k+\eta_k\bm u_k}}.
    \label{eq:principle_wn_update}
\end{align}
WD shrinks \(\bm\theta_k\) by \(1-\lambda\eta_k\) before adding \(\eta_k\bm u_k\), allowing the block's norm \(r_k:=\norm{\bm\theta_k}\) to evolve.
WN instead projects \(\bm\theta_k+\eta_k\bm u_k\) back to norm \(R\), keeping \(r_k\equiv R\) while preserving direction.

In modern LLMs with post-norm~\citep{vaswani2017attention} or pre-norm~\citep{xiong2020prenorm} layers inducing scale-invariance, optimization depends primarily on how each update changes the parameter direction, which scales inversely with the block's norm \(r_k\).
We therefore define the blockwise norm-based ELR~\citep{vanlaarhoven2017l2,dangelo2024weightdecay,kosson2026wdtransfer} at step \(k\) as
\begin{equation}
    \elr_k:=\frac{\eta_k}{r_k}.
    \label{eq:principle_elr}
\end{equation}
Each block has its own scalar \(\elr_k\) at step \(k\) and traces an ELR trajectory \(\{\elr_k\}_{k\geq 0}\) across training.
For small relative steps, it represents the scale of directional change (Appendices~\ref{app:new_normalized_updates} and~\ref{app:new_tangential_geometry}).

\subsection{Aligning the training dynamics of WN and WD via ELR matching}
\label{sec:elr_principle_alignment}

\textbf{Experimental setup.}
To validate that matching the blockwise ELR in Eq.~\eqref{eq:principle_elr} aligns the training dynamics of WN and WD, we match their ELR trajectories in both directions, either matching WN to WD's ELR trajectory or matching WD to WN's in the reverse direction.
Here an \emph{ELR target} is the complete trajectory of per-step sequences \(\{\elr_k\}_{k\geq 0}\).
For example, to make WN match WD's ELR under cosine decay, we first record each block's per-step parameter norm \(\{r_k^{\mathrm{WD}}\}_{k\geq 0}\) from a standard WD run to obtain its blockwise ELR trajectory \(\{\elr_k\}_{k\geq 0}\).
Then, the matching can be done in two ways:
(i) Via \emph{norm replay}.
We keep the nominal LR \(\eta_k^{\mathrm{WN}}=\eta_k^{\mathrm{WD}}\) unchanged and set each WN block's target norm at step \(k\) to the recorded \(r_k^{\mathrm{WD}}\).
(ii) Via \emph{LR scaling}.
We keep the standard WN operation, normalizing each WN block to its fixed norm \(R\), and rescale its nominal LR at step \(k\) to satisfy
\begin{equation}
    \frac{\eta_k^{\mathrm{WN}}}{R}
    =
    \frac{\eta_k^{\mathrm{WD}}}{r_k^{\mathrm{WD}}}
    \implies
    \eta_k^{\mathrm{WN}}
    =R\frac{\eta_k^{\mathrm{WD}}}{r_k^{\mathrm{WD}}}.
    \label{eq:elr_transfer}
\end{equation}

\begin{figure}[t]
    \centering
    \includegraphics[width=\textwidth]{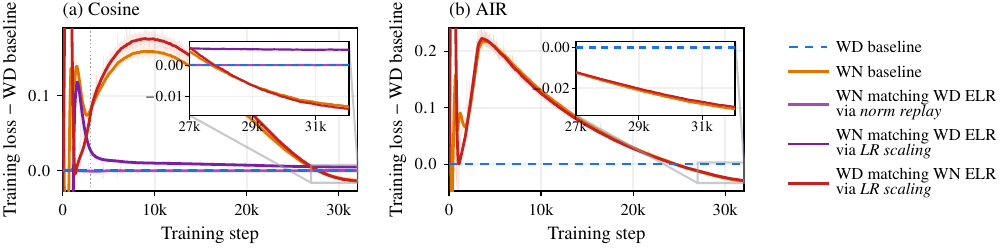}
    \vspace{ - 15pt}
    \caption{\textbf{Matching blockwise ELR trajectories collapses the loss dynamics of WN and WD.}
    Each curve shows per-step training loss minus the corresponding WD baseline.
    Matching WN to WD's ELR pulls the curve onto WD, and vice versa across both schedules.}
    \label{fig:elr_dynamics_collapse}
\end{figure}

In the reverse direction, we make WD match WN's ELR trajectory \(\{\eta_k^{\mathrm{WN}}/R\}_{k\geq 0}\) via \emph{LR scaling}.
Because changing the nominal LR in WD would also alter subsequent parameter norms, we replay the recorded WD norms \(r_k^{\mathrm{WD}}\) at each step \(k\) and rescale each block's nominal LR to \(\eta_k^{\mathrm{WD}}=(r_k^{\mathrm{WD}}/R)\eta_k^{\mathrm{WN}}\). Details are given in Appendix~\ref{app:new_elr_transfer}.
We evaluate this reverse matching under both cosine decay (Figure~\ref{fig:elr_dynamics_collapse}(a)) and \OurMethod, the schedule derived in Section~\ref{sec:wn_schedule} (Figure~\ref{fig:elr_dynamics_collapse}(b)).

\textbf{Results.}
Figure~\ref{fig:elr_dynamics_collapse} shows that matching blockwise ELR trajectories collapses the loss dynamics of WN and WD onto nearly the same curve across both directions and schedules.
Under cosine decay (Figure~\ref{fig:elr_dynamics_collapse}(a)), matching WN to WD's ELR via \emph{norm replay} nearly reproduces the WD baseline with a terminal gap below \(2\times 10^{-4}\), while matching via \emph{LR scaling} also closely tracks WD, leaving a small residual of several \(\times 10^{-3}\).
In the reverse direction, matching WD to WN's ELR via \emph{LR scaling} closely tracks the WN baseline under both cosine decay and \OurMethod (Figure~\ref{fig:elr_dynamics_collapse}(a,b)).
Together, these results confirm that the ELR trajectory accounts for most of the loss dynamics difference between WD and WN, making ELR a principled target for schedule design.

\subsection{WN as a clean setting for ELR schedule design}
\label{sec:elr_principle_design}

Under WD, changing the nominal LR \(\eta_k\) also alters subsequent parameter norms \(r_{>k}\), coupling the two factors in Eq.~\eqref{eq:principle_elr} so that rescaling \(\eta_k\) alone cannot prescribe a target ELR trajectory~\citep{kosson2026wdtransfer}.
WN removes this coupling by fixing \(r_k\equiv R\), giving \(\elr_k=\eta_k/R\), making each block's ELR directly proportional to its nominal LR~\citep{wen2026hyperball,an2026demystifyingmanifoldconstraintsllm}.
This makes WN a clean setting where ELR scheduling reduces to designing the nominal LR schedule.
At the same time, our matching results in Section~\ref{sec:elr_principle_alignment} provide the foundation for this WN-based derivation, demonstrating that an ELR trajectory derived under WN can be transferred to WD via ELR matching to reproduce its loss dynamics (Figure~\ref{fig:elr_dynamics_collapse}(b)).
Therefore, we derive our ELR schedule under WN, starting from its update rule in Section~\ref{sec:wn_schedule}.

\section{Designing a Better ELR Schedule under WN: \OurMethod}
\label{sec:wn_schedule}


\subsection{Directional-alignment dynamics under WN}
\label{sec:wn_schedule_dynamics}

\textbf{From the WN update to directional increments.}
We start from the WN parameter update in Eq.~\eqref{eq:principle_wn_update}, which fixes \(\norm{\bm\theta_k}=R\) and updates only the unit direction \(\hat{\bm\theta}_k=\bm\theta_k/R\).
Decompose each update \(\bm u_k\) into radial and tangential components relative to \(\hat{\bm\theta}_k\):
\begin{equation}
    \bm u_k=u_{\parallel,k}\hat{\bm\theta}_k+\bm u_k^\perp,
    \qquad
    u_{\parallel,k}:=\hat{\bm\theta}_k^\top\bm u_k,
    \qquad
    \hat{\bm\theta}_k^\top\bm u_k^\perp=0.
    \label{eq:wn_schedule_decomposition}
\end{equation}
Substituting Eq.~\eqref{eq:wn_schedule_decomposition} into Eq.~\eqref{eq:principle_wn_update} gives the exact unit-direction update:
\begin{equation}
    \hat{\bm\theta}_{k+1}
    =\frac{\bm\theta_k+\eta_k\bm u_k}{\norm{\bm\theta_k+\eta_k\bm u_k}}
    =\frac{(R+\eta_k u_{\parallel,k})\hat{\bm\theta}_k+\eta_k\bm u_k^\perp}
    {\sqrt{(R+\eta_k u_{\parallel,k})^2+\eta_k^2\norm{\bm u_k^\perp}^2}}.
    \label{eq:wn_schedule_tangential_identity}
\end{equation}
Expanding to \(O(\eta_k^2)\) for small relative steps (\(\eta_k\norm{\bm u_k}/R\ll 1\)) yields (derivation in Appendix~\ref{app:wn_schedule_increment})
\begin{equation}
    \hat{\bm\theta}_{k+1}-\hat{\bm\theta}_k
    =\frac{\eta_k}{R}\!\left(1-\frac{\eta_k u_{\parallel,k}}{R}\right)\bm u_k^\perp
    -\frac{\eta_k^2\norm{\bm u_k^\perp}^2}{2R^2}\hat{\bm\theta}_k
    +O(\eta_k^3).
    \label{eq:wn_schedule_increment}
\end{equation}

\textbf{Alignment dynamics.}
Let \(\bm v\) denote the optimal unit parameter direction on the sphere.
Conditioned on the decay-start history \(\hist_0\), define the expected alignment with \(\bm v\), tangential drift, and retraction curvature:
\begin{equation}
    m_k:=\E\!\left[\bm v^\top\hat{\bm\theta}_k\middle|\hist_0\right],
    \qquad
    a_k:=\E\!\left[\frac{\bm v^\top\bm u_k^\perp}{R}\middle|\hist_0\right]>0,
    \qquad
    b_k:=\frac{\E\!\left[\norm{\bm u_k^\perp}^2\middle|\hist_0\right]}{2R^2}>0,
    \label{eq:wn_schedule_alignment}
\end{equation}
averaging over minibatch randomness across the first \(k\) steps from \(\hist_0\).
Projecting Eq.~\eqref{eq:wn_schedule_increment} onto \(\bm v\) and taking \(\E[\cdot\mid\hist_0]\) yields the discrete alignment update and its continuous-time ODE in decay time \(t\geq 0\) (detailed derivation in Appendix~\ref{app:wn_schedule_dynamics}):
\begin{align}
    m_{k+1}-m_k &= a_k\eta_k-b_k\eta_k^2 m_k+O(\eta_k^3),
    \label{eq:wn_schedule_recurrence}\\
    \dot m(t) &= a(t)\eta(t)-b(t)\eta(t)^2 m(t),
    \qquad
    m(0)=m_0>0.
    \label{eq:wn_schedule_ode}
\end{align}
A larger LR boosts new alignment linearly while strengthening spherical pullback of existing alignment quadratically.
Solving Eq.~\eqref{eq:wn_schedule_ode} on \([0,t]\) via integrating factors gives (detailed derivation in Appendix~\ref{app:wn_schedule_integral})
\begin{equation}
    m(t)=m_0 H(t,0)+\int_0^t H(t,s)a(s)\eta(s)\,\dd s,
    \qquad
    H(t,s):=\exp\!\left[-\int_s^t b(u)\eta(u)^2\,\dd u\right],
    \label{eq:wn_schedule_integral}
\end{equation}
where \(H(t,s)\in(0,1]\) is the alignment fraction from time \(s\) retained at \(t\).

\subsection{Optimal schedule without endpoint constraints: inverse-square-root}
\label{sec:wn_schedule_optimal}

\textbf{General optimum via completing the square.}
To maximize \(m(t)\) at every \(t\geq 0\) without prescribed peak, terminal, or cumulative LRs, multiply Eq.~\eqref{eq:wn_schedule_ode} by \(2m(t)\) and complete the square in \(\eta(t)\):
\begin{equation}
    \frac{\dd}{\dd t}m(t)^2
    =\frac{a(t)^2}{2b(t)}-2b(t)\left(m(t)\eta(t)-\frac{a(t)}{2b(t)}\right)^2
    \leq\frac{a(t)^2}{2b(t)}.
    \label{eq:wn_schedule_square}
\end{equation}
Integrating over \([0,t]\) shows that \(m(t)\) attains its upper bound for all \(t\geq 0\) if and only if the squared term vanishes almost everywhere (proof in Appendix~\ref{app:wn_schedule_optimal}), yielding the general optimum:
\begin{equation}
    m^\star(t)=\sqrt{m_0^2+\tau(t)},
    \qquad
    \eta^\star(t)=\frac{a(t)}{2b(t)\sqrt{m_0^2+\tau(t)}},
    \qquad
    \tau(t):=\int_0^t\frac{a(s)^2}{2b(s)}\,\dd s.
    \label{eq:wn_schedule_general_opt}
\end{equation}

\begin{figure}[t]
    \centering
    \includegraphics[width=\textwidth]{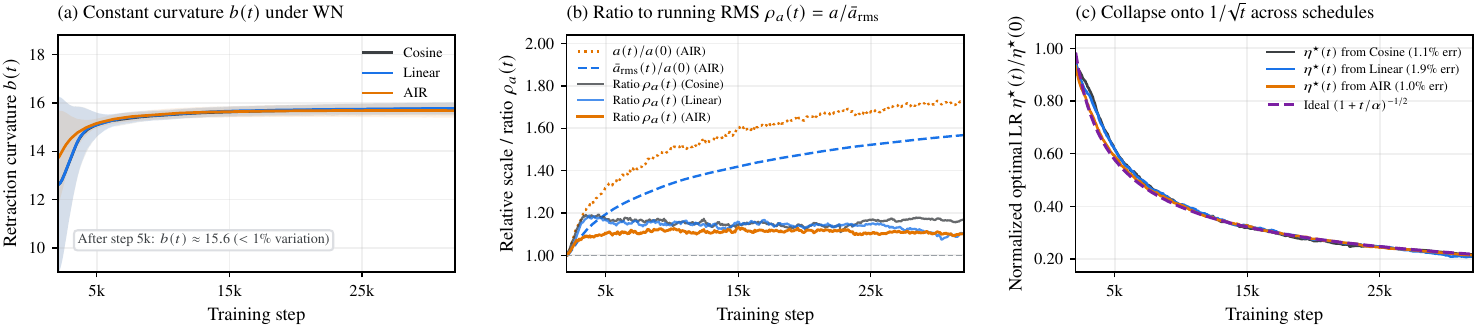}
    \vspace{-14pt}
    \caption{\textbf{MoE pretraining under WN empirically satisfies the conditions for $1/\sqrt{t}$ decay across schedules.}
    Post-warmup blockwise statistics confirm nearly constant curvature \(b(t)\) and a stable ratio \(\rho_a(t)=a(t)/\bar a_{\mathrm{rms}}(t)\), so substituting the measured \((a(t),b(t))\) into Eq.~\eqref{eq:wn_schedule_general_opt} makes the optimal schedule \(\eta^\star(t)/\eta^\star(0)\) closely match \((1+t/\alpha)^{-1/2}\).}
    \label{fig:sec3_self_normalization}
    \vspace{ - 12pt}
\end{figure}

\textbf{From the general optimum to \(1/\sqrt{t}\) decay.}
To simplify Eq.~\eqref{eq:wn_schedule_general_opt} without requiring constant \(a(t)\), define the running root-mean-square (RMS) of \(a(t)\) on \([0,t]\) and the ratio of \(a(t)\) to its running RMS:
\begin{equation}
    \bar a_{\mathrm{rms}}(t):=\sqrt{\frac{1}{t}\int_0^t a(s)^2\,\dd s},
    \qquad
    \rho_a(t):=\frac{a(t)}{\bar a_{\mathrm{rms}}(t)}.
    \label{eq:wn_schedule_rms_def}
\end{equation}
Empirically under WN (Figure~\ref{fig:sec3_self_normalization}), (i) \(b(t)\equiv b\) is constant, so \(\tau(t)=\bar a_{\mathrm{rms}}(t)^2 t/(2b)\), and (ii) \(a(t)\) is non-vanishing and slowly varying with nearly constant \(\rho_a(t)\approx\rho>0\); together, these reduce Eq.~\eqref{eq:wn_schedule_general_opt} directly to shifted \(1/\sqrt{t}\) decay (Appendix~\ref{app:wn_schedule_selfnorm}):
\begin{equation}
    \eta^\star(t)
    =\frac{\eta_0}{\sqrt{\left(\frac{a(0)}{a(t)}\right)^2+\frac{2b\eta_0^2}{\rho_a(t)^2}\,t}}
    \approx\frac{\eta_0}{\sqrt{1+t/\alpha}},
    \qquad
    \eta_0:=\frac{a(0)}{2bm_0},\quad \alpha:=\frac{\rho^2}{2b\eta_0^2}.
    \label{eq:wn_schedule_optimal_lr}
\end{equation}
Whereas \(1/\sqrt{t}\) decay is classical in horizon-free nominal schedules~\citep{shamir2013sgd,vaswani2017attention,meterez2026anytime}, arises passively without WD from growing norms~\citep{kosson2026wdtransfer}, or couples LR with WD~\citep{su2025ema}, here it emerges directly from maximizing directional alignment under WN.

\textbf{Empirical verification.}
To verify conditions (i) and (ii) above, we track blockwise curvature \(b(t)\) and drift \(a(t)\) during MoE pretraining under WN across Cosine, Linear, and \OurMethod (measurement protocol in Appendix~\ref{app:wn_schedule_protocol}).
As shown in Figure~\ref{fig:sec3_self_normalization}, \(b(t)\) stabilizes around \(15.6\) (\(<1\%\) variation; panel~\textbf{(a)}), while \(a(t)\) rises without vanishing and \(\rho_a(t)\) remains flat within \(1.07\text{--}1.18\) (panel~\textbf{(b)}).
Substituting the measured \((a(t),b(t))\) directly into Eq.~\eqref{eq:wn_schedule_general_opt} collapses all three optimal schedules \(\eta^\star(t)/\eta^\star(0)\) onto \((1+t/\alpha)^{-1/2}\) within \(1.0\%\text{--}1.9\%\) median error (panel~\textbf{(c)}), demonstrating that the conditions underlying Eq.~\eqref{eq:wn_schedule_optimal_lr} hold in practice.

\subsection{\OurMethod: Finite-horizon schedule design via affine calibration}
\label{sec:wn_schedule_hart}

\textbf{Why raw \(1/\sqrt{t}\) requires finite-horizon annealing.}
Section~\ref{sec:wn_schedule_optimal} derives \(1/\sqrt{t}\) decay without prescribed endpoints, whereas practical pretraining decays over \(K\) steps from peak LR \(p\) to a near-zero floor \(\eta_{\min}\) (e.g., $10^{-5}$).
Under our measured statistics in Eq.~\eqref{eq:wn_schedule_optimal_lr}, setting \(\eta_0=p\) places \(\alpha\) on the order of \(10^3\text{--}10^4\) steps, leaving raw \(1/\sqrt{t}\) decay near \(p/3\gg\eta_{\min}\) at step \(K\) and sustaining large noise at the end of training.
Prior \(1/\sqrt{t}\) schedules likewise require finite-horizon annealing, either implicitly via iterate or weight averaging~\citep{shamir2013sgd,meterez2026anytime} or explicitly via piecewise cooldown~\citep{ma2026wsqd}.
Meaningful comparison therefore requires an annealed setting. Following standard pretraining practice, we compare schedules under the same terminal LR \(\eta_{\min}\) without weight averaging.

\textbf{Affine schedule design (\OurMethod).}
To decay smoothly from \(\eta(0)=p\) to \(\eta(K)=\eta_{\min}\) in a single closed-form curve, we apply an affine transformation to \(1/\sqrt{t}\) decay, yielding \textbf{\OurMethod} (Appendix~\ref{app:wn_schedule_affine_endpoints}):
\begin{gather}
    q(k)=\left(1+\frac{k}{\alpha}\right)^{-1/2},
    \qquad
    f(k)=\frac{q(k)-q(K)}{1-q(K)},
    \label{eq:wn_schedule_shape}\\
    \eta_{\OurMethodMath}(k)
    =\eta_{\min}+(p-\eta_{\min})f(k),
    \qquad
    k=0,\ldots,K.
    \label{eq:wn_schedule_hart}
\end{gather}
We set \(\alpha=K/8\) by default (evaluating sensitivity to \(\alpha/K\) in Section~\ref{sec:new_evaluation_robustness}) and calibrate \(p\) from the full discrete LR sum, including warmup, when matching cumulative LR (Appendix~\ref{app:wn_schedule_lr_sum}).

\textbf{Advantages of the affine design.}
Although enforcing \(\eta(K)=\eta_{\min}\) departs from the free-endpoint optimum in Eq.~\eqref{eq:wn_schedule_optimal_lr}, among smooth schedules matching \(\eta(0)=p\) and \(\eta(K)=\eta_{\min}\), this affine map uniquely preserves both decay-rate ratios across \(t_1,t_2\in[0,K]\) and the fraction of total LR reduction over every interval \(0\leq i<j\leq K\):
\begin{equation}
    \frac{\dot\eta_{\OurMethodMath}(t_1)}{\dot\eta_{\OurMethodMath}(t_2)}
    =\frac{\dot q(t_1)}{\dot q(t_2)},
    \qquad
    \frac{\eta_{\OurMethodMath}(i)-\eta_{\OurMethodMath}(j)}{p-\eta_{\min}}
    =\frac{q(i)-q(j)}{1-q(K)}.
    \label{eq:wn_schedule_affine_decrements}
\end{equation}
Scaling all decay rates by a common positive factor minimizes relative decay-rate distortion (Appendix~\ref{app:wn_schedule_affine_endpoints}).
It also keeps the schedule strictly convex and smooth without extra cooldown hyperparameters.
Moreover, under convex last-iterate bounds~\citep{defazio2024linear,schaipp2025agreement}, \(1/\sqrt{t}\) is the unique decay family whose bound-minimizing finite-horizon schedule reduces exactly to this affine form (Appendix~\ref{app:wn_schedule_tailsum}).
Finally, applying the same affine calibration across all candidate decay shapes in Section~\ref{sec:new_evaluation} isolates the empirical benefit of \(q(k)\) itself.

\section{Empirical Evaluation}
\label{sec:new_evaluation}

\subsection{Experimental setup}
\label{sec:new_evaluation_setup}

\textbf{Pretraining setting.}
We pretrain a mixture-of-experts (MoE) model with 110\,M activated parameters
and 4.7\,B total parameters on 128\,B tokens using Muon~\citep{jordan2024muon} under WN.
Each complete run takes 48 hours on 16 GPUs.
To further test scalability, we apply the same scheduler design to a larger MoE
with 790\,M activated and 35.7\,B total parameters, trained on 500\,B tokens,
with each run taking 168 hours on 32 GPUs.
Additional experimental settings and reporting conventions are given in Appendix~\ref{app:new_protocol}.

\textbf{Comparison controls.}
To ensure an optimal baseline, we search the peak LR under the WN cosine baseline and select the optimal peak LR of \(1.07\times 10^{-3}\), alongside the warmup ratio (Figure~\ref{fig:cosine_baseline_lr_sweep} in Appendix~\ref{app:new_protocol_controls}).
To compare how different decay shapes distribute the same total LR budget across training, all WN schedule families share the same linear warmup duration, terminal LR, and cumulative LR, defined as the sum of LRs over all training steps (peak calibration in Appendix~\ref{app:wn_schedule_lr_sum}).
All schedules are compared at the end of training.

\begin{figure}[t]
    \centering
    \begin{minipage}[t]{0.59\linewidth}
        \vspace{0pt}
        \centering
        \includegraphics[width=\linewidth]{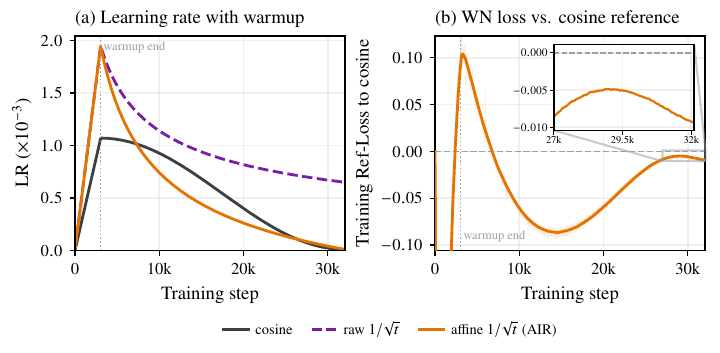}
        \vspace{-14pt}
        \caption{\textbf{Learning rate schedules and training loss under WN.}
        With the same terminal and cumulative LRs, \OurMethod achieves lower
        final training loss than cosine.}
        \label{fig:hart_warmup_lr_refloss}
    \end{minipage}\hfill
    \begin{minipage}[t]{0.38\linewidth}
        \vspace{0pt}
        \centering
        \includegraphics[width=\linewidth]{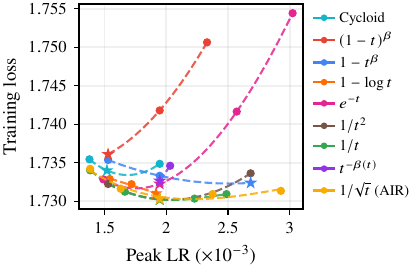}
        \vspace{ - 18pt}
        \caption{\textbf{\OurMethod stays near its best loss over the widest
        peak LR range.} Circles are measured configurations, stars the best
        per family, and dashed curves family-wise convex fits.}
        \label{fig:lr_family_training_loss}
    \end{minipage}
    \vspace{ - 12pt}
\end{figure}

Figure~\ref{fig:hart_warmup_lr_refloss}(a) shows the cosine baseline,
raw \(1/\sqrt{t}\) decay~\citep{vaswani2017attention}, and its affine transformation \OurMethod,
defined by Eqs.~\eqref{eq:wn_schedule_shape}--\eqref{eq:wn_schedule_hart}.
The raw curve is shown as a reference with the same peak as \OurMethod,
without matching its terminal or cumulative LR.
We also compare other decay shapes under the same controls and affine
construction, including \(e^{-t}\)~\citep{xiao2026hyperball},
\(1-\log(1+t)\)~\citep{loshchilov2026ngpt}, and
\(1-\sqrt{t}\)~\citep{hagele2024cooldown,zhou-zhou-gu-2026-elr}.
Appendix~\ref{app:new_protocol_shapes} lists the complete set of decay shapes.

\subsection{Pretraining loss and downstream performance}
\label{sec:new_evaluation_performance}

\textbf{Experimental setup.}
Each WN decay family is evaluated at its best shape parameter (Table~\ref{tab:comparison_decay_shapes}) on terminal losses and 11 downstream benchmarks (Appendix~\ref{app:protocol_downstream_benchmarks}), alongside WD cosine baselines at both model scales.
Additional WD cosine and WSD comparisons with ELR trajectories are in Appendix~\ref{app:protocol_existing_recipes}.

\textbf{Results.}
Figure~\ref{fig:hart_warmup_lr_refloss}(b) and Table~\ref{tab:multi_scheduler_compare} show that \OurMethod lowers terminal training and evaluation loss by about 0.01 relative to WN cosine at both scales, achieving the lowest losses among all compared schedules and substantially outperforming WD cosine.
\OurMethod also attains the highest downstream average across all shapes at 128\,B tokens and improves 10 of 11 benchmarks over WN cosine at 500\,B tokens with +4.05 average points over WD cosine baseline, demonstrating strong scalability.

\begin{table*}[t]
\caption{\textbf{Terminal losses and downstream performance of matched-budget decay settings.}
Cosine is reported under both weight decay (WD) and weight normalization (WN); all other decay shapes use WN.
Each WN shape is labeled by its raw decay shape before the common affine transformation and uses that shape's best parameter selected during evaluation (Appendix~\ref{app:new_protocol_shapes}).
\OurMethod achieves the lowest terminal losses and attains the best downstream average at both model scales.}
\label{tab:multi_scheduler_compare}
\begin{center}
\begin{adjustbox}{max width=\textwidth}
{\setlength{\tabcolsep}{3.5pt}%
\begin{tabular}{lcc|ccccccccccc|c}
    \toprule
    \begin{tabular}[c]{@{}l@{}}Setting\end{tabular} & \begin{tabular}[c]{@{}c@{}}Train\\Loss ($\downarrow$)\end{tabular} & \begin{tabular}[c]{@{}c@{}}Eval\\Loss ($\downarrow$)\end{tabular} & \begin{tabular}[c]{@{}c@{}}MMLU\\($\uparrow$)\end{tabular} & \begin{tabular}[c]{@{}c@{}}MMLU-Pro\\($\uparrow$)\end{tabular} & \begin{tabular}[c]{@{}c@{}}AGIEval\\($\uparrow$)\end{tabular} & \begin{tabular}[c]{@{}c@{}}WinoGrande\\($\uparrow$)\end{tabular} & \begin{tabular}[c]{@{}c@{}}BBH\\($\uparrow$)\end{tabular} & \begin{tabular}[c]{@{}c@{}}DROP\\($\uparrow$)\end{tabular} & \begin{tabular}[c]{@{}c@{}}GSM8k\\($\uparrow$)\end{tabular} & \begin{tabular}[c]{@{}c@{}}MATH\\($\uparrow$)\end{tabular} & \begin{tabular}[c]{@{}c@{}}HumanEval\\($\uparrow$)\end{tabular} & \begin{tabular}[c]{@{}c@{}}MBPP+\\($\uparrow$)\end{tabular} & \begin{tabular}[c]{@{}c@{}}TriviaQA\\($\uparrow$)\end{tabular} & \begin{tabular}[c]{@{}c@{}}Avg.\\($\uparrow$)\end{tabular} \\
    \midrule
    \multicolumn{15}{l}{\textbf{110\,M/4.7\,B model}} \\
    \addlinespace[2pt]
    Cosine (WD) & 1.753 & 2.205 & 46.7 & 16.0 & 25.9 & 50.5 & 30.8 & 21.7 & 13.6 & 11.2 & 26.2 & 35.4 & 30.3 & 28.03 \\
    Cosine (WN) & 1.740 & 2.185 & 49.0 & 18.2 & 29.1 & 49.5 & 30.9 & 22.0 & 17.0 & 11.1 & 28.7 & 35.2 & 30.9 & 29.24 \\
    Linear & 1.742 & 2.187 & 48.9 & 16.4 & 26.3 & 49.6 & 31.1 & 22.6 & 13.4 & 11.3 & 26.2 & 35.4 & 31.2 & 28.40 \\
    \addlinespace
    Half-cosine & 1.749 & 2.195 & 48.6 & 15.5 & 28.2 & 50.7 & \textbf{32.5} & \textbf{23.9} & 15.0 & 11.7 & 24.4 & 34.7 & 29.5 & 28.61 \\
    Rear-cosine & 1.736 & 2.180 & 48.0 & 16.2 & 28.7 & \textbf{51.2} & 31.6 & 22.3 & 14.3 & 10.5 & 25.0 & 36.2 & 30.7 & 28.61 \\
    Cycloid & 1.734 & 2.178 & 47.3 & 16.0 & 27.5 & 49.9 & 30.7 & 22.6 & 15.1 & 10.0 & 26.2 & 33.6 & 31.1 & 28.18 \\
    \addlinespace
    $(1-t)^2$ & 1.736 & 2.181 & 48.1 & 16.8 & 27.6 & 50.7 & 30.6 & 22.1 & 16.5 & 12.0 & 27.4 & \textbf{36.8} & 29.5 & 28.92 \\
    $1-t^{\beta}$ & 1.732 & \textbf{2.174} & 48.5 & 18.1 & \textbf{30.2} & 50.9 & 31.8 & 22.3 & 16.4 & 10.9 & \textbf{31.1} & 35.2 & 32.3 & 29.79 \\
    \addlinespace
    $1-\log t$ & 1.731 & 2.175 & 48.1 & 15.6 & 29.6 & 48.8 & 32.0 & 21.5 & 16.8 & 11.6 & 23.8 & 34.4 & \textbf{33.2} & 28.67 \\
    $e^{-t}$ & 1.732 & 2.176 & 48.9 & 16.0 & 29.2 & 50.6 & 31.6 & 21.2 & 16.8 & 10.8 & 27.4 & \textbf{36.8} & 31.7 & 29.18 \\
    $1/t^2$ & \textbf{1.730} & 2.175 & \textbf{50.5} & 18.4 & 29.0 & 50.8 & 32.1 & 22.2 & \textbf{17.5} & 11.6 & 29.9 & 33.1 & 30.3 & 29.58 \\
    $1/t$ & \textbf{1.730} & \textbf{2.174} & 49.7 & 16.9 & 29.6 & 50.8 & 30.8 & 23.0 & 16.7 & 12.2 & 26.8 & 32.8 & 29.7 & 29.00 \\
    $t^{-\beta(t)}$ & 1.733 & 2.176 & 48.8 & 16.7 & 30.1 & 50.7 & 30.0 & 22.9 & 14.2 & 10.6 & 28.0 & 34.7 & 30.6 & 28.85 \\
    \rowcolor{gray!15} $1/\sqrt{t}$ (\OurMethod) & \textbf{1.730} & \textbf{2.174} & 49.7 & \textbf{18.6} & 29.6 & 50.1 & 31.1 & 23.8 & 15.8 & \textbf{13.0} & 29.9 & 36.5 & 31.6 & \textbf{29.97} \\
    \midrule
    \multicolumn{15}{l}{\textbf{790\,M/35.7\,B model}} \\
    \addlinespace[2pt]
    Cosine (WD) & 1.423 & 1.857 & 68.2 & 37.9 & 52.9 & 50.9 & 52.5 & 47.5 & 60.3 & 33.7 & 67.1 & 55.3 & 64.7 & 53.73 \\
    Cosine (WN) & 1.431 & 1.843 & 70.7 & 40.6 & 54.2 & 50.4 & 56.2 & \textbf{56.0} & 59.9 & 36.8 & 67.7 & 60.1 & 65.8 & 56.22 \\
    \rowcolor{gray!15} $1/\sqrt{t}$ (\OurMethod) & \textbf{1.421} & \textbf{1.832} & \textbf{72.1} & \textbf{42.4} & \textbf{55.7} & \textbf{51.8} & \textbf{59.4} & 51.2 & \textbf{64.2} & \textbf{37.2} & \textbf{70.1} & \textbf{65.3} & \textbf{66.2} & \textbf{57.78} \\
    \bottomrule
  \end{tabular}%
}%

\end{adjustbox}
\end{center}
\end{table*}

\subsection{Robustness to hyperparameter choices}
\label{sec:new_evaluation_robustness}

\textbf{Experimental setup.}
Under the same matched controls, we sweep each family's shape parameter, which also moves
its peak LR. For each family we report its LR sensitivity \(S\), the average loss above the
family's best configuration per e-fold of peak LR, defined formally in
Appendix~\ref{app:new_protocol_downstream} (Eq.~\eqref{eq:lr_sensitivity}).

\textbf{Results.}
Figure~\ref{fig:lr_family_training_loss} plots terminal training loss against peak LR with
family-wise polynomial fits, and Table~\ref{tab:lr_family_robustness} in
Appendix~\ref{app:new_protocol_downstream} reports the resulting
LR sensitivity.
\OurMethod has the lowest LR sensitivity over the widest range of peak LR.
Its region of low loss is therefore wide enough that the gain does not depend on a narrowly
tuned peak LR and thus facilitates
LR transfer~\citep{ren2026hyperp}.

\section{Late-Stage Scheduling Matters More}
\label{sec:new_late_stage}


\subsection{Late-stage scheduling drives the final loss gap}
\label{sec:new_late_attribution}

\begin{figure}[t]
    \centering
    \includegraphics[width=\linewidth]{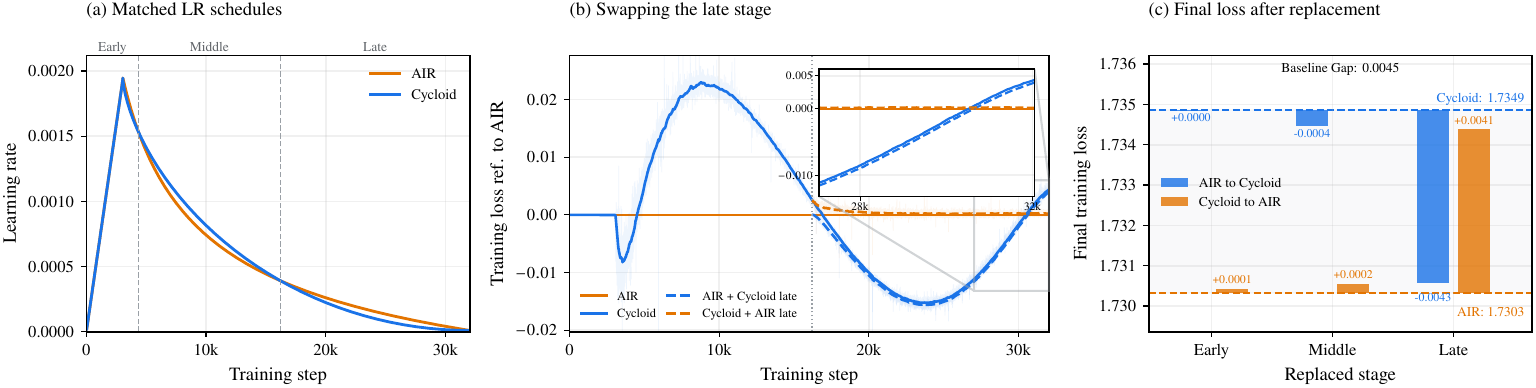}
    \caption{\textbf{Late-stage scheduling dominates the final loss
    difference.}
    Starting from two full schedules with matched peak, terminal, and cumulative LRs, swapping the late stage
    between \OurMethod and Cycloid nearly transfers the full final loss gap to
    either schedule, whereas swaps of the early and middle stages have
    little effect.
    The bars use the terminal exponential moving average (EMA-0.99) training loss, as in Table~\ref{tab:multi_scheduler_compare}; the intervention and reporting details are given in Appendix~\ref{app:protocol_stage_swaps}.}
    \label{fig:fmr_stage_summary}
\end{figure}

Among the candidate schedules matched to the same
cumulative LR, a Cycloid schedule with $\theta_{1} = \pi$ happens to take
exactly the same peak LR as \OurMethod, so the two runs share their warmup and
terminal LR as well and differ only in the shape of the decay in between
(Figure~\ref{fig:fmr_stage_summary}(a)). 
However, their final training losses still
differ by approximately 0.005, a substantial gap at this scale. 
Because the two intersections of their decay curves naturally divide
the training horizon into early, middle, and late stages, we swap
one stage at a time between them while leaving the other two unchanged.

As shown in Figure~\ref{fig:fmr_stage_summary}(b,c), swaps of the early and middle stages
leave the final loss nearly unchanged, whereas swapping the late stage
closes about 90\,\% of the gap for either schedule.
For this matched pair, the gap is almost entirely attributable to the late stage.
This motivates us to examine whether a shared late schedule can reduce the final loss gap created by distinct early schedules.
A related effect was reported for scale-invariant networks trained with weight decay, where reaching the equilibrium of a late phase makes the outcome largely independent of earlier phases~\citep{li2020intrinsic}.

\subsection{Shared late scheduling reduces the final loss gap}
\label{sec:new_late_shared}

\begin{table*}[htbp]
    \caption{\textbf{A shared late schedule narrows the endpoint gap for every
        early decay shape.}
        Each schedule is trained twice, once in its original form and once with the
        common late schedule. Parentheses give the gap of each loss to the
        \OurMethod anchor in the last row, and \emph{Gap closed} is the percentage
        of that gap removed by the shared tail.
        Training loss is the terminal EMA-0.99 value and evaluation loss the
        terminal raw value, as in Table~\ref{tab:multi_scheduler_compare}.}
        \label{tab:fixrear_own_vs_shared_tail}
    \vspace{ - 4pt}
    \begin{center}
\definecolor{gapgreen}{RGB}{110,150,110}%
\definecolor{gaporig}{RGB}{176,42,42}%
\definecolor{gapshared}{RGB}{31,116,64}%
\begin{adjustbox}{max width=\textwidth}%
{\setlength{\tabcolsep}{3.5pt}%
\begin{tabular}{lcccccc}
    \toprule
    \multirow{2}{*}{Schedule} & \multicolumn{3}{c}{Training loss {\scriptsize(relative to \OurMethod)}} & \multicolumn{3}{c}{Eval loss {\scriptsize(relative to \OurMethod)}} \\
    \cmidrule(lr){2-4}\cmidrule(lr){5-7}
    & Original & Shared tail & Gap closed (\%) $\uparrow$ & Original & Shared tail & Gap closed (\%) $\uparrow$ \\
    \midrule
    Cosine & 1.7399\,{\scriptsize\textcolor{gaporig}{$(+0.0096)$}} & 1.7358\,{\scriptsize\textcolor{gapshared}{$(+0.0055)$}} & \cellcolor{gapgreen!12!white}42\% & 2.1846\,{\scriptsize\textcolor{gaporig}{$(+0.0106)$}} & 2.1799\,{\scriptsize\textcolor{gapshared}{$(+0.0059)$}} & \cellcolor{gapgreen!14!white}45\% \\
    Linear & 1.7415\,{\scriptsize\textcolor{gaporig}{$(+0.0112)$}} & 1.7341\,{\scriptsize\textcolor{gapshared}{$(+0.0038)$}} & \cellcolor{gapgreen!38!white}66\% & 2.1875\,{\scriptsize\textcolor{gaporig}{$(+0.0135)$}} & 2.1787\,{\scriptsize\textcolor{gapshared}{$(+0.0047)$}} & \cellcolor{gapgreen!38!white}65\% \\
    Half-cosine & 1.7487\,{\scriptsize\textcolor{gaporig}{$(+0.0184)$}} & 1.7374\,{\scriptsize\textcolor{gapshared}{$(+0.0071)$}} & \cellcolor{gapgreen!34!white}62\% & 2.1950\,{\scriptsize\textcolor{gaporig}{$(+0.0210)$}} & 2.1815\,{\scriptsize\textcolor{gapshared}{$(+0.0076)$}} & \cellcolor{gapgreen!36!white}64\% \\
    Double-cosine & 1.7446\,{\scriptsize\textcolor{gaporig}{$(+0.0143)$}} & 1.7328\,{\scriptsize\textcolor{gapshared}{$(+0.0025)$}} & \cellcolor{gapgreen!57!white}83\% & 2.1901\,{\scriptsize\textcolor{gaporig}{$(+0.0161)$}} & 2.1769\,{\scriptsize\textcolor{gapshared}{$(+0.0029)$}} & \cellcolor{gapgreen!57!white}82\% \\
    $(1-t)^2$ & 1.7361\,{\scriptsize\textcolor{gaporig}{$(+0.0058)$}} & 1.7319\,{\scriptsize\textcolor{gapshared}{$(+0.0016)$}} & \cellcolor{gapgreen!46!white}72\% & 2.1805\,{\scriptsize\textcolor{gaporig}{$(+0.0065)$}} & 2.1753\,{\scriptsize\textcolor{gapshared}{$(+0.0013)$}} & \cellcolor{gapgreen!54!white}80\% \\
    $1-\sqrt{t}$ & 1.7354\,{\scriptsize\textcolor{gaporig}{$(+0.0050)$}} & 1.7316\,{\scriptsize\textcolor{gapshared}{$(+0.0012)$}} & \cellcolor{gapgreen!49!white}75\% & 2.1787\,{\scriptsize\textcolor{gaporig}{$(+0.0047)$}} & 2.1743\,{\scriptsize\textcolor{gapshared}{$(+0.0003)$}} & \cellcolor{gapgreen!70!white}94\% \\
    \midrule
    $1/\sqrt{t}$ (\OurMethod) & 1.7303\,{\scriptsize\phantom{$(+0.0000)$}} & -- & -- & 2.1740\,{\scriptsize\phantom{$(+0.0000)$}} & -- & -- \\
    \bottomrule
\end{tabular}%
}%
\end{adjustbox}%

        \end{center}
    \end{table*}
    
\textbf{Experimental setup.}
To address this question, we consider six decay schedules, Cosine, Linear, Half-cosine, Double-cosine, \((1-t)^2\), and \(1-\sqrt{t}\), whose original runs end with training losses 0.005--0.018 higher than \OurMethod.
For each schedule, we retain its decay shape over the first 70\,\% of training steps and concatenate it with the same late-stage schedule of \OurMethod.
The early segment and its warmup target are rescaled to match the cumulative LR under the same comparison constraints as in Section~\ref{sec:new_evaluation_setup}.
The two runs of each schedule are denoted \emph{original} and \emph{shared tail}, respectively.

\textbf{Results.}
Table~\ref{tab:fixrear_own_vs_shared_tail} shows that the shared late schedule lowers terminal training and evaluation loss for all six schedules, closing 42\,\%--83\,\% and 45\,\%--94\,\% of their respective gaps to \OurMethod.
Thus, a shared late schedule substantially reduces the performance differences arising from different early schedules.

\subsection{A trade-off in late-stage scheduling}
\label{sec:new_late_exponents}

\textbf{Experimental setup.}
Having established the importance of late-stage scheduling, we next examine what makes a late schedule effective.
Based on the design of \OurMethod in Section~\ref{sec:wn_schedule_hart}, we vary the final 20\,\% of its decay stage, starting at \(K_s\approx0.8K\), as
\begin{equation}
    \eta_\gamma(k)=\eta_{\min}
    +\bigl[\eta_{\OurMethodMath}(K_s)-\eta_{\min}\bigr]
    \left(\frac{K-k}{K-K_s}\right)^\gamma,
    \quad K_s\leq k\leq K,
    \label{eq:new_late_gamma}
\end{equation}
where \(\gamma\in\{0.5,0.75,1,2,3\}\).
Smaller exponents postpone more of the LR reduction to the endpoint, whereas larger exponents reduce the LR earlier and then flatten out.
The junction and terminal LRs remain fixed, while the area under the recorded LR traces varies by less than 1\,\% (Appendix~\ref{app:protocol_late_exponents}).

\begin{figure}[h!]
    \centering
    \includegraphics[width=\linewidth]{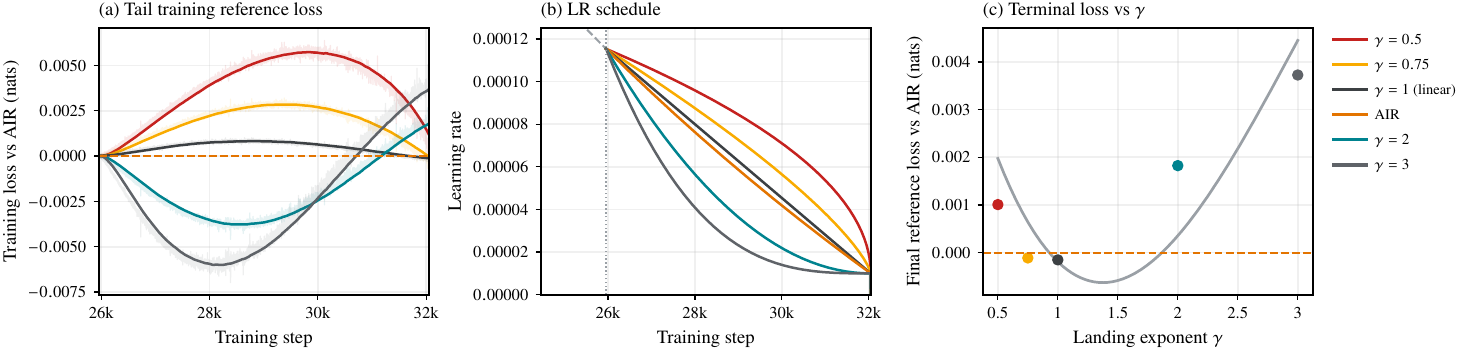}
    \caption{\textbf{Late-stage decay exhibits a two-sided trade-off.}
    With the junction and terminal LRs fixed, both overly abrupt decay and early flattening increase final loss.}
    \label{fig:tail_landing_refloss}
\end{figure}

\textbf{Results.}
Figure~\ref{fig:tail_landing_refloss} shows that both the abrupt decay at \(\gamma=0.5\) and the early flattening at \(\gamma\in\{2,3\}\) increase final loss, while \OurMethod and the variants with \(\gamma\in\{0.75,1\}\) perform comparably, achieving the lowest final losses.
These results suggest a trade-off: late-stage scheduling should sustain annealing while avoiding overly abrupt LR changes.
The design of \OurMethod naturally balances these requirements and performs close to the best variants in this sweep.

\subsection{A theoretical explanation of schedule dependence}
\label{sec:new_late_theory}

To interpret these results, we use the alignment dynamics of Section~\ref{sec:wn_schedule} to separate the effects of early- and late-stage schedule changes.
Consider bounded nonnegative schedules \(\eta_1,\eta_2\) on the decay interval \([0,K]\), with common bounded trajectory \(a(k)>0\), constant curvature \(b>0\) (Section~\ref{sec:wn_schedule_optimal}), initial state \(m_0,\hist_0\), and stage boundary \(K_s\).
Write \(\Delta m=m_1-m_2\) and \(\Delta\eta=\eta_1-\eta_2\), where the expected alignments \(m_i\) retain the conditioning in Eq.~\eqref{eq:wn_schedule_alignment} throughout.
For a shared late schedule, \(\eta_1=\eta_2=\eta\) on \([K_s,K]\), Eq.~\eqref{eq:wn_schedule_integral} gives
\begin{equation}
    \Delta m(K)=H(K,K_s)\Delta m(K_s),\quad
    H(K,K_s)=\exp\!\left[-b\int_{K_s}^K\eta(k)^2\,\dd k\right].
    \label{eq:new_common_tail}
\end{equation}
Since \(0<H(K,K_s)\leq1\), the shared late stage attenuates the entering difference without reversing its sign, ignoring stochastic fluctuations around the mean.
Over a fixed late interval, higher LRs increase the cumulative squared LR and reduce \(H(K,K_s)\), thereby weakening the effect of early-stage scheduling on endpoint alignment.
The derivation is given in Appendix~\ref{app:comparison_common_tail}.

For different late-stage schedules, a shared early schedule gives \(\Delta m(K_s)=0\).
For bounds \(0\leq\eta_i\leq P\) and \(0\leq m_i\leq M\) on \([K_s,K]\), let \(L_{12}=\sup_{k\in[K_s,K]}\max\{a(k),|a(k)-2bPM|\}\) and \(H_{12}=\sqrt{H_1H_2}\), with \(H_i\) the retention factor for \(\eta_i\).
This gives
\begin{equation}
    |\Delta m(K)|\leq L_{12}\int_{K_s}^K
    H_{12}(K,k)|\Delta\eta(k)|\,\dd k.
    \label{eq:new_tail_distance}
\end{equation}
Smaller weighted LR differences give a tighter bound on the terminal alignment difference.
Within this bound, differences nearer the endpoint receive greater weight because less subsequent attenuation remains.
Details are given in Appendix~\ref{app:comparison_different_tails}.
Further, if the schedules differ in both stages,
\begin{equation}
    |\Delta m(K)|\leq H_{12}(K,K_s)|\Delta m(K_s)|
    +L_{12}\int_{K_s}^K H_{12}(K,k)|\Delta\eta(k)|\,\dd k.
    \label{eq:new_full_schedule_bound}
\end{equation}
The first term retains the early-stage difference, while the second bounds the late-stage contribution.
Slope-based refinements are in Appendix~\ref{app:comparison_slopes}.
These bounds on expected alignment provide a qualitative explanation for the observed loss gaps.

\section{Related Work}
\label{sec:new_related_work}

\textbf{Weight-normalized training.}
WN and related constraints explicitly control parameter scale as an alternative to WD, with reported efficiency, stability, and transfer gains~\citep{wen2026hyperball,an2026demystifyingmanifoldconstraintsllm}.
Methods differ in whether they separate magnitude from direction~\citep{salimans2016wn}, normalize weight vectors and representations onto the hypersphere~\citep{loshchilov2025ngpt}, fix matrix-weight and update norms~\citep{wen2026hyperball}, explore alternative matrix norms~\citep{xie2026sso,xu2026moga}, or retract tangent-space descent steps~\citep{kexuefm-11196,kexuefm-11221,bernstein2025manifolds}.
However, while \citet{wen2026hyperball} motivate fixed norms through effective step-size control, these works adopt standard nominal LR schedules without examining ELR alignment between WN and WD.

\textbf{Learning-rate scheduling.}
Scheduling theories explain WSD via river-valley geometry~\citep{wen2025wsd} and motivate cooldown and power decay via convex bounds and functional scaling laws~\citep{schaipp2025agreement,li2026optimal}, while prior $1/\sqrt{t}$ schedules study horizon-free nominal LRs in non-smooth convex SGD~\citep{shamir2013sgd} or anytime pretraining with dynamic weight averaging~\citep{meterez2026anytime}.
Yet these works focus on nominal LR alone, which is not principled when norms evolve.
Extending \citeposs{dangelo2024weightdecay} approach of matching the \emph{global} ELR of WD in unregularized training, we perform bidirectional \emph{blockwise} alignment between WN and WD to motivate ELR schedule design.

\textbf{Concurrent work.}
Concurrent studies likewise corroborate that matching ELR aligns WN and WD loss dynamics.
In particular, they study angular step-size attribution~\citep{xiao2026hyperball}, the scope of ELR collapse and ELR-based functional scaling laws~\citep{liu2026elr}, angular-update heuristics and empirical schedules~\citep{zhou-zhou-gu-2026-elr}, or low-cost pretraining recipes~\citep{luo2026puro2b}, yet none derives an optimal ELR schedule.
We instead derive $1/\sqrt{t}$ ELR decay directly from directional-alignment dynamics under WN and adapt it via an affine transformation into {\OurMethod} for finite-horizon pretraining without weight averaging.

\section{Conclusion and Future Work}
\label{sec:new_conclusion}

The effective learning rate (ELR) trajectory accounts for most of the observed loss dynamics difference between WN and WD, and matching ELR substantially narrows their training loss gap.
This finding motivates schedule design that targets ELR directly.
Modeling directional-alignment dynamics under WN yields $1/\sqrt{t}$ decay when no peak or terminal LR is prescribed, and an affine adaptation gives \OurMethod for finite-horizon pretraining.
Experiments on MoE models at two scales under WN show that, at matched cumulative LR, \OurMethod reduces terminal loss by about 0.01 relative to cosine and improves downstream performance, while schedule sweeps demonstrate hyperparameter robustness.
Further stage-wise interventions show that sharing the late schedule substantially narrows final loss gaps from different early schedules, identifying late-stage scheduling as a key source of \OurMethod's effectiveness.
Transferring \OurMethod's ELR trajectory to WD reproduces the corresponding WN training loss dynamics, demonstrating that our design extends beyond fixed-norm training.

Future work includes quantitatively connecting directional alignment to training loss and incorporating stochastic noise into finite-horizon analysis.
We also aim to investigate the ELR principle and \OurMethod across more architectures, optimizers, and larger training scales, and to study ELR scheduling in continued training (CT), supervised fine-tuning (SFT), and reinforcement learning (RL).

\label{newmaintext:end}
\clearpage

\bibliographystyle{plainnat}
\bibliography{references}

\clearpage
\beginappendix
\section{Directional Geometry and ELR Matching}
\label{app:new_geometry}

We derive the geometric interpretation of ELR in Section~\ref{sec:elr_principle_updates} and the blockwise matching rule used in Section~\ref{sec:elr_principle_alignment}.

\subsection{Exact normalized updates}
\label{app:new_normalized_updates}

For a nonzero parameter block, retain \(r_k=\norm{\bm\theta_k}\) and \(\elr_k=\eta_k/r_k\) from Section~\ref{sec:elr_principle_updates}, and write \(\hat{\bm\theta}_k=\bm\theta_k/r_k\) for its unit direction.
Throughout, \(\eta_k\geq0\), and norms are Euclidean on vectorized blocks.
Under fixed-norm WN, \(r_k=R\).
Dividing the updates in Eqs.~\eqref{eq:principle_wd_update}
and~\eqref{eq:principle_wn_update} by their pre-update parameter norms and
normalizing gives
\begin{align}
    \text{WD:}\quad
    \hat{\bm\theta}_{k+1}
    &=\frac{(1-\lambda\eta_k)\hat{\bm\theta}_k+\elr_k\bm u_k}
    {\norm{(1-\lambda\eta_k)\hat{\bm\theta}_k+\elr_k\bm u_k}},
    \label{eq:new_wd_direction}\\
    \text{WN:}\quad
    \hat{\bm\theta}_{k+1}
    &=\frac{\hat{\bm\theta}_k+\elr_k\bm u_k}
    {\norm{\hat{\bm\theta}_k+\elr_k\bm u_k}},\qquad
    \elr_k=\eta_k/R.
    \label{eq:new_wn_direction}
\end{align}
These identities hold whenever the vector being normalized is nonzero.
They keep the
realized optimizer update explicit. The ELR multiplies that update relative
to the current parameter norm, while WD also contributes radial shrinkage.

\subsection{Radial and tangential components}
\label{app:new_tangential_geometry}

Define
\[
    u_{\parallel,k}=\hat{\bm\theta}_k^\top\bm u_k,
    \qquad
    \bm u_k^\perp=\bm u_k-u_{\parallel,k}\hat{\bm\theta}_k.
\]
The scalar \(u_{\parallel,k}\) is the radial component, and
\(\bm u_k^\perp\) is the component orthogonal to the parameter direction.
Under WD, the updated vector is
\[
    \bigl[(1-\lambda\eta_k)r_k+\eta_k u_{\parallel,k}\bigr]
    \hat{\bm\theta}_k+\eta_k\bm u_k^\perp.
\]
Orthogonality gives its squared norm as the sum of the squared radial and tangential terms.
The unsigned angle \(\phi_k\in[0,\pi]\) between the old and updated directions is therefore
\begin{equation}
    \phi_k=\operatorname{atan2}\!\left(
    \eta_k\norm{\bm u_k^\perp},
    (1-\lambda\eta_k)r_k+\eta_k u_{\parallel,k}\right).
    \label{eq:new_exact_angle}
\end{equation}
For WN, the same expression applies with \(\lambda=0\) and \(r_k=R\),
because the final normalization preserves direction.

To obtain the first-order change, the derivative of
\(\bm x\mapsto\bm x/\norm{\bm x}\) at a unit vector
\(\hat{\bm\theta}_k\) is
\(\bm I-\hat{\bm\theta}_k\hat{\bm\theta}_k^\top\).
Applied to the increment
\(-\lambda\eta_k\hat{\bm\theta}_k+\elr_k\bm u_k\), it removes the radial
term and gives
\begin{equation}
    \hat{\bm\theta}_{k+1}-\hat{\bm\theta}_k
    =\elr_k\bm u_k^\perp
    +O\!\left((\lambda\eta_k+\elr_k\norm{\bm u_k})^2\right).
    \label{eq:new_first_order_direction}
\end{equation}
The expansion requires \(\lambda\eta_k+\elr_k\norm{\bm u_k}\ll1\).
The WN expression again sets \(\lambda=0\).
For a given tangential optimizer update, the ELR thus sets the first-order scale of the directional change.

\subsection{Blockwise ELR transfer}
\label{app:new_elr_transfer}

For a target trajectory \(\eta_k^{\mathrm{eff,tar}}\), choosing
\(\eta_k^{\mathrm{WD}}=r_k^{\mathrm{WD}}\eta_k^{\mathrm{eff,tar}}\)
gives \(\eta_k^{\mathrm{WD}}/r_k^{\mathrm{WD}}=\eta_k^{\mathrm{eff,tar}}\)
by substitution.
A WN schedule supplies the target \(\eta_k^{\mathrm{eff,tar}}=\eta_k^{\mathrm{WN}}/R\), giving
\begin{equation}
    \eta_k^{\mathrm{WD}}=\frac{r_k^{\mathrm{WD}}}{R}\eta_k^{\mathrm{WN}}.
    \label{eq:new_elr_transfer}
\end{equation}
This is the reverse direction of the coefficient matching in Eq.~\eqref{eq:elr_transfer}.
We apply this rule separately to each controlled block, using the recorded WD norm \(r_k^{\mathrm{WD}}\) replayed at step \(k\) in the reverse-matching experiments.

\section{WN Schedule Design and Calibration}
\label{app:wn_schedule}

The subsections below follow the exact order in which they are referenced in Section~\ref{sec:wn_schedule} (Appendices~\ref{app:wn_schedule_increment}--\ref{app:wn_schedule_tailsum}), with each subsection proving one step of the main-text derivation.

\subsection{Second-order directional increment}
\label{app:wn_schedule_increment}

We derive the exact unit-direction update in Eq.~\eqref{eq:wn_schedule_tangential_identity} and its second-order expansion in Eq.~\eqref{eq:wn_schedule_increment} from the WN parameter update Eq.~\eqref{eq:principle_wn_update}.
Under fixed-norm WN (\(\norm{\bm\theta_k}=R>0\)), write \(\bm\theta_k=R\hat{\bm\theta}_k\) with \(\norm{\hat{\bm\theta}_k}=1\).
Substituting the orthogonal decomposition \(\bm u_k=u_{\parallel,k}\hat{\bm\theta}_k+\bm u_k^\perp\) from Eq.~\eqref{eq:wn_schedule_decomposition} gives the pre-normalized vector and, by \(\hat{\bm\theta}_k^\top\bm u_k^\perp=0\), its Euclidean norm:
\[
    \bm\theta_k+\eta_k\bm u_k
    =(R+\eta_k u_{\parallel,k})\hat{\bm\theta}_k+\eta_k\bm u_k^\perp,
    \qquad
    \norm{\bm\theta_k+\eta_k\bm u_k}
    =\sqrt{(R+\eta_k u_{\parallel,k})^2+\eta_k^2\norm{\bm u_k^\perp}^2}.
\]
Dividing Eq.~\eqref{eq:principle_wn_update} by \(R\) and substituting both expressions yields Eq.~\eqref{eq:wn_schedule_tangential_identity}.

For small relative steps \(\eta_k\norm{\bm u_k}/R\ll 1\), we have \(\eta_k|u_{\parallel,k}|<R\) and hence \(R+\eta_k u_{\parallel,k}>0\).
Dividing both numerator and denominator of Eq.~\eqref{eq:wn_schedule_tangential_identity} by \(R+\eta_k u_{\parallel,k}\) separates the scalar normalization factor from the tangential step:
\[
    \hat{\bm\theta}_{k+1}
    =\left(1+\frac{\eta_k^2\norm{\bm u_k^\perp}^2}{(R+\eta_k u_{\parallel,k})^2}\right)^{\!-1/2}
    \!\left(\hat{\bm\theta}_k+\frac{\eta_k}{R+\eta_k u_{\parallel,k}}\bm u_k^\perp\right).
\]
Using \((1+x)^{-1}=1-x+O(x^2)\) and \((1+x)^{-1/2}=1-x/2+O(x^2)\) as \(\eta_k\norm{\bm u_k}/R\to 0\), the two factors expand up to \(O(\eta_k^2)\) as
\[
    \begin{aligned}
        \frac{\eta_k}{R+\eta_k u_{\parallel,k}}
        &=\frac{\eta_k}{R}\!\left(1+\frac{\eta_k u_{\parallel,k}}{R}\right)^{\!-1}
        =\frac{\eta_k}{R}\!\left(1-\frac{\eta_k u_{\parallel,k}}{R}\right)+O(\eta_k^3),\\
        \left(1+\frac{\eta_k^2\norm{\bm u_k^\perp}^2}{(R+\eta_k u_{\parallel,k})^2}\right)^{\!-1/2}
        &=1-\frac{\eta_k^2\norm{\bm u_k^\perp}^2}{2(R+\eta_k u_{\parallel,k})^2}+O(\eta_k^4)
        =1-\frac{\eta_k^2\norm{\bm u_k^\perp}^2}{2R^2}+O(\eta_k^3),
    \end{aligned}
\]
where \(u_{\parallel,k}\) in the denominator of the second line enters only at \(O(\eta_k^3)\).
Multiplying these two expansions and subtracting \(\hat{\bm\theta}_k\) yields Eq.~\eqref{eq:wn_schedule_increment}:
\[
    \begin{aligned}
        \hat{\bm\theta}_{k+1}-\hat{\bm\theta}_k
        &=\left(1-\frac{\eta_k^2\norm{\bm u_k^\perp}^2}{2R^2}\right)
        \!\left[\hat{\bm\theta}_k+\frac{\eta_k}{R}\!\left(1-\frac{\eta_k u_{\parallel,k}}{R}\right)\bm u_k^\perp\right]
        -\hat{\bm\theta}_k+O(\eta_k^3)\\
        &=\frac{\eta_k}{R}\!\left(1-\frac{\eta_k u_{\parallel,k}}{R}\right)\bm u_k^\perp
        -\frac{\eta_k^2\norm{\bm u_k^\perp}^2}{2R^2}\hat{\bm\theta}_k
        +O(\eta_k^3).
    \end{aligned}
\]

\subsection{Discrete alignment update and continuous-time ODE}
\label{app:wn_schedule_dynamics}

\textbf{Discrete alignment update (Eq.~\eqref{eq:wn_schedule_recurrence}).}
Taking the inner product of Eq.~\eqref{eq:wn_schedule_increment} with \(\bm v\) and applying \(\E[\cdot\mid\hist_0]\) for a deterministic schedule \(\{\eta_k\}\) gives, with \(m_k,a_k,b_k\) in Eq.~\eqref{eq:wn_schedule_alignment},
\begin{equation}
    m_{k+1}-m_k
    =a_k\eta_k-b_k\eta_k^2 m_k
    -\eta_k^2\left(
        \mathrm{Cov}\!\left(\frac{\norm{\bm u_k^\perp}^2}{2R^2},\bm v^\top\hat{\bm\theta}_k\middle|\hist_0\right)
        +\E\!\left[\frac{u_{\parallel,k}\,\bm v^\top\bm u_k^\perp}{R^2}\middle|\hist_0\right]
    \right)
    +O(\eta_k^3).
    \label{eq:wn_schedule_expansion}
\end{equation}
When relative tangential step norms concentrate around their RMS scale \(\sqrt{2b_k}\) (as under normalized updates such as Muon, making the covariance negligible) and the radial--tangential cross term is suppressed relative to \(a_k\eta_k\) by \(\eta_k|u_{\parallel,k}|/R\ll 1\), Eq.~\eqref{eq:wn_schedule_expansion} reduces directly to the discrete alignment update Eq.~\eqref{eq:wn_schedule_recurrence}.

\textbf{Mutual derivation between discrete and continuous dynamics.}
In the small-step regime \(\sqrt{2b_k}\eta_k\ll 1\), the discrete update in Eq.~\eqref{eq:wn_schedule_recurrence} and the continuous ODE in Eq.~\eqref{eq:wn_schedule_ode} imply each other up to \(O(\eta_k^3)\):
\begin{enumerate}
    \item \emph{Continuous to discrete (exact one-step integration).}
    Suppose \(m(t)\) satisfies Eq.~\eqref{eq:wn_schedule_ode} with step values \((a_k, b_k, \eta_k)\) on \(t\in[k, k+1]\).
    Multiplying by \(e^{b_k\eta_k^2(t-k)}\) and integrating over \([k, k+1]\) gives the exact one-step map
    \[
        m(k+1)
        =e^{-b_k\eta_k^2}m(k)+a_k\eta_k\,\frac{1-e^{-b_k\eta_k^2}}{b_k\eta_k^2}.
    \]
    Expanding \(e^{-x}=1-x+O(x^2)\) and \((1-e^{-x})/x=1-x/2+O(x^2)\) at \(x=b_k\eta_k^2=O(\eta_k^2)\) yields
    \[
        m(k+1)-m(k)
        =a_k\eta_k-b_k\eta_k^2 m(k)-\frac{1}{2}a_k b_k\eta_k^3+O(\eta_k^4)
        =a_k\eta_k-b_k\eta_k^2 m(k)+O(\eta_k^3),
    \]
    recovering the discrete update in Eq.~\eqref{eq:wn_schedule_recurrence}.

    \item \emph{Discrete to continuous (small-step continuum limit).}
    Conversely, let smooth trajectories \(m(t)\), \(a(t)\), \(b(t)\), and \(\eta(t)\) interpolate \((m_k, a_k, b_k, \eta_k)\) at \(t=k\) while evolving on the macroscopic horizon scale \(\Theta(\eta^{-2})\) (so \(\dot\eta=O(\eta^3)\) and \(\dot a, \dot b=O(\eta^2)\)).
    By Taylor's theorem on \([k, k+1]\), substituting Eq.~\eqref{eq:wn_schedule_recurrence} gives
    \[
        \dot m(k)+\frac{1}{2}\ddot m(k)+O(\dddot m)
        =m(k+1)-m(k)
        =a(k)\eta(k)-b(k)\eta(k)^2 m(k)+O(\eta_k^3).
    \]
    Differentiating the leading relation \(\dot m=a\eta-b\eta^2 m+O(\eta^3)\) gives
    \[
        \ddot m=\dot a\eta+a\dot\eta-b\eta^2\dot m+O(\eta^4)=O(\eta^3),
    \]
    so matching terms through \(O(\eta_k^2)\) yields the continuous ODE in Eq.~\eqref{eq:wn_schedule_ode}.
\end{enumerate}

\subsection{Continuous and discrete retention solutions}
\label{app:wn_schedule_integral}

\textbf{Continuous integral solution (Eq.~\eqref{eq:wn_schedule_integral}).}
To solve Eq.~\eqref{eq:wn_schedule_ode} on \([0,t]\) with \(m(0)=m_0>0\), take bounded positive \(a(s),b(s)>0\) and nonnegative \(\eta\in L^2([0,t])\).
Multiplying the linear ODE at time \(s\) by its integrating factor gives
\[
    \frac{\dd}{\dd s}\left(
    m(s)\exp\!\left[\int_0^s b(u)\eta(u)^2\,\dd u\right]\right)
    =a(s)\eta(s)\exp\!\left[\int_0^s b(u)\eta(u)^2\,\dd u\right].
\]
Integrating from \(s=0\) to \(s=t\) and dividing by the terminal factor weights each contribution from time \(s\) by the retention ratio
\[
    \frac{\exp\!\left[\int_0^s b(u)\eta(u)^2\,\dd u\right]}
         {\exp\!\left[\int_0^t b(u)\eta(u)^2\,\dd u\right]}
    =\exp\!\left[-\int_s^t b(u)\eta(u)^2\,\dd u\right]
    =H(t,s),
\]
yielding
\begin{equation}
    m(t)=m_0 H(t,0)+\int_0^t H(t,s)a(s)\eta(s)\,\dd s,
    \label{eq:wn_schedule_integral_proof}
\end{equation}
which proves Eq.~\eqref{eq:wn_schedule_integral} and ensures \(m(t)>0\) for all \(t\geq 0\).

\textbf{Discrete retention counterpart.}
Iterating \(m_{j+1}=(1-b_j\eta_j^2)m_j+a_j\eta_j+O(\eta_j^3)\) from Eq.~\eqref{eq:wn_schedule_recurrence} across \(j=0,\ldots,k-1\) yields
\[
    m_k=m_0 H_{k,0}+\sum_{j=0}^{k-1}H_{k,j+1}a_j\eta_j+O\!\left(\sum_{j=0}^{k-1}\eta_j^3\right),
    \qquad
    H_{k,i}:=\prod_{s=i}^{k-1}\bigl(1-b_s\eta_s^2\bigr),
\]
where \(\ln H_{k,i}=-\sum_{s=i}^{k-1}b_s\eta_s^2+O(\sum_{s=i}^{k-1}\eta_s^4)\) matches \(\ln H(t,s)\) up to second order.

\subsection{General optimal schedule}
\label{app:wn_schedule_optimal}

For any nonnegative \(\eta\in L^2([0,t])\) and positive \(a(s),b(s)>0\), integrating Eq.~\eqref{eq:wn_schedule_square} over \([0,t]\) with \(\tau(t)\) and \(m^\star(t)\) defined in Eq.~\eqref{eq:wn_schedule_general_opt} yields the exact alignment-gap identity
\begin{equation}
    m^\star(t)^2-m(t)^2
    =\int_0^t 2b(s)\left(m(s)\eta(s)-\frac{a(s)}{2b(s)}\right)^2\dd s\geq 0.
    \label{eq:wn_schedule_optimal_gap}
\end{equation}
Thus \(m(t)\leq m^\star(t)\) for every \(t\geq 0\), with equality at all \(t\geq 0\) if and only if \(m(t)\eta(t)=a(t)/[2b(t)]\) almost everywhere.
Under this condition, \(\dd(m^2)/\dd t=a(t)^2/[2b(t)]\), so \(m(t)=m^\star(t)\) and \(\eta^\star(t)=a(t)/[2b(t)m^\star(t)]\), proving Eq.~\eqref{eq:wn_schedule_general_opt}.

\subsection{Reduction to inverse-square-root decay under non-vanishing, slowly varying drift}
\label{app:wn_schedule_selfnorm}

\textbf{Exact denominator identity.}
Given \(b(t)\equiv b\), we have \(\tau(t)=\int_0^t \frac{a(s)^2}{2b}\,\dd s=\frac{\bar a_{\mathrm{rms}}(t)^2}{2b}\,t\) for \(\bar a_{\mathrm{rms}}(t)\) in Eq.~\eqref{eq:wn_schedule_rms_def}.
Writing the initial value as \(\eta_0:=\eta^\star(0)=\frac{a(0)}{2bm_0}\) (so \(2bm_0=a(0)/\eta_0\)) and dividing both numerator and denominator of Eq.~\eqref{eq:wn_schedule_general_opt} by \(a(t)/\eta_0\) establishes the exact equality in Eq.~\eqref{eq:wn_schedule_optimal_lr}:
\[
    \eta^\star(t)
    =\frac{a(t)}{2b\sqrt{m_0^2+\frac{\bar a_{\mathrm{rms}}(t)^2}{2b}\,t}}
    =\frac{\eta_0}{\sqrt{\left(\frac{2bm_0\eta_0}{a(t)}\right)^2+\frac{2b\eta_0^2\bar a_{\mathrm{rms}}(t)^2}{a(t)^2}\,t}}
    =\frac{\eta_0}{\sqrt{\left(\frac{a(0)}{a(t)}\right)^2+\frac{2b\eta_0^2}{\rho_a(t)^2}\,t}}.
\]

\textbf{Log-slope identity and reduction to \((1+t/\alpha)^{-1/2}\).}
For any absolutely continuous \(a(t)>0\) with log-slope \(\varepsilon_a(s):=s\dot a(s)/a(s)\), integration by parts gives
\[
    \int_0^t a(s)^2\,\dd s
    =t\,a(t)^2-2\int_0^t s\,a(s)\dot a(s)\,\dd s
    =t\,a(t)^2-2\int_0^t \varepsilon_a(s)a(s)^2\,\dd s.
\]
Dividing both sides by \(\int_0^t a(s)^2\,\dd s=t\,\bar a_{\mathrm{rms}}(t)^2\) expresses \(\rho_a(t)\) in terms of the \(a^2\)-weighted average log-slope \(\bar\varepsilon_a(t)\):
\[
    \rho_a(t)=\frac{a(t)}{\bar a_{\mathrm{rms}}(t)}=\sqrt{1+2\bar\varepsilon_a(t)},
    \qquad
    \bar\varepsilon_a(t):=\frac{\int_0^t \varepsilon_a(s)a(s)^2\,\dd s}{\int_0^t a(s)^2\,\dd s}.
\]
When \(a(t)\) varies slowly so that \(\bar\varepsilon_a(t)>-1/2\) stays nearly constant, \(\rho_a(t)\approx\rho>0\) stays nearly constant, making the \(t\)-coefficient \(\frac{2b\eta_0^2}{\rho_a(t)^2}\approx\frac{2b\eta_0^2}{\rho^2}=:\frac{1}{\alpha}\) nearly constant.
To compare the exact radicand in Eq.~\eqref{eq:wn_schedule_optimal_lr} against \(1+t/\alpha\) over a finite horizon where \(t/\alpha\) may be moderate (e.g., \(K/\alpha=8\)) rather than asymptotically infinite, write their difference as
\[
    \left[\left(\frac{a(0)}{a(t)}\right)^2+\frac{2b\eta_0^2}{\rho_a(t)^2}\,t\right]-\left(1+\frac{t}{\alpha}\right)
    =\left[\left(\frac{a(0)}{a(t)}\right)^2-1\right]+\left(\frac{\rho^2}{\rho_a(t)^2}-1\right)\frac{t}{\alpha}.
\]
Three properties preserve the shifted inverse-square-root form without requiring \(1\) to be negligible relative to \(t/\alpha\):
(i) \(\bigl(a(0)/a(t)\bigr)^2\) equals \(1\) at \(t=0\) (staying near \(1\) at small \(t\) by slow variation), and non-vanishing \(a(t)\geq a_{\min}>0\) keeps \(\bigl(a(0)/a(t)\bigr)^2\in(0,a(0)^2/a_{\min}^2]\) bounded of order one (specifically \(\in(0,1]\) when \(a(t)\geq a(0)\) in Figure~\ref{fig:sec3_self_normalization}(b)), preventing it from blowing up to compete with the growing \(t/\alpha\) term;
(ii) when \(a(t)\) drifts monotonically from \(a(0)\), \(\bigl(a(0)/a(t)\bigr)^2-1\) and \(\bigl(\rho^2/\rho_a(t)^2-1\bigr)(t/\alpha)\) take opposite signs and partially offset each other; and
(iii) even if \(\bigl(a(0)/a(t)\bigr)^2\) settles near a bounded level \(c>0\) (\(c\neq 1\)) at moderate \(t/\alpha\), \(\eta_0/\sqrt{c+t/\alpha}=(\eta_0/\sqrt{c})/\sqrt{1+t/(c\alpha)}\) remains an exact shifted inverse-square-root profile with rescaled constants \((\eta_0/\sqrt{c},\,c\alpha)\), while at fixed \((\eta_0,\alpha)\) the square root dampens any bounded radicand discrepancy \(|\Delta R|\) by \(2(1+t/\alpha)\) in relative LR.

\textbf{Exact all-time cancellation for shifted power laws.}
In particular, for any shifted power law \(a(t)=a_0(1+t/\alpha)^\beta\) (\(\beta>-1/2\)) with \(m_0^2=\frac{a_0^2\alpha}{2b(1+2\beta)}\) (so \(\lim_{t\to\infty}\rho_a(t)=\rho=\sqrt{1+2\beta}\) and \(2b\eta_0^2=\frac{1+2\beta}{\alpha}\)), direct integration gives \(\tau(t)=m_0^2\bigl[(1+t/\alpha)^{1+2\beta}-1\bigr]\), and the transient deviations of the two terms inside the square root cancel identically at every \(t\geq 0\):
\[
    \left(\frac{a(0)}{a(t)}\right)^2+\frac{2b\eta_0^2}{\rho_a(t)^2}\,t
    =\left(1+\frac{t}{\alpha}\right)^{\!-2\beta}
    +\left[1+\frac{t}{\alpha}-\left(1+\frac{t}{\alpha}\right)^{\!-2\beta}\right]
    =1+\frac{t}{\alpha},
\]
so \(\eta^\star(t)=\eta_0/\sqrt{1+t/\alpha}\) holds identically for all \(t\geq 0\).

\subsection{Empirical tracking protocol for Figure~\ref{fig:sec3_self_normalization}}
\label{app:wn_schedule_protocol}

In the 110\,M-activated MoE pretraining runs under WN (Section~\ref{sec:new_evaluation_setup}) with Cosine, Linear, and \OurMethod schedules, we log blockwise statistics after warmup across the \(32{,}047\)-step schedule (from global step \(k_{\mathrm{warm}}=3{,}040\) to \(32{,}046\), indexed by decay time \(t=k-k_{\mathrm{warm}}\geq 0\)), smoothed with EMA-\(0.9\) and averaged across layers within each linear module type before taking the mean (and cross-module standard deviation) across module types.

\textbf{Exact decomposition of \(a_k\) and \(b_k\).}
Let \(\bm g_k\) denote the parameter block's minibatch gradient at step \(k\), with tangential component \(\bm g_k^\perp:=\bm g_k-(\hat{\bm\theta}_k^\top\bm g_k)\hat{\bm\theta}_k\), and write \(\cos(\bm x,\bm y):=\bm x^\top\bm y/(\norm{\bm x}\norm{\bm y})\) for any nonzero vectors \(\bm x,\bm y\).
From Eq.~\eqref{eq:wn_schedule_decomposition}, \(\norm{\bm u_k^\perp}^2=\norm{\bm u_k}^2[1-\cos^2(\bm\theta_k,\bm u_k)]\), so \(b(t)\) at \(t=k-k_{\mathrm{warm}}\) is determined directly from logged norms and cosines:
\begin{equation}
    b(t)
    =\frac{\norm{\bm u_k^\perp}^2}{2R^2}
    =\frac{\norm{\bm u_k}^2}{2R^2}\bigl[1-\cos^2(\bm\theta_k,\bm u_k)\bigr].
    \label{eq:wn_schedule_measured_b}
\end{equation}
For \(a_k\) in Eq.~\eqref{eq:wn_schedule_alignment}, expanding \(\bm v^\top\bm u_k^\perp=\bm v^\top\bm u_k-(\bm v^\top\hat{\bm\theta}_k)(\hat{\bm\theta}_k^\top\bm u_k)\) with \(\norm{\bm v}=1\) and using Eq.~\eqref{eq:wn_schedule_measured_b} gives the exact identity
\begin{equation}
    \frac{\bm v^\top\bm u_k^\perp}{R}
    =\sqrt{2b(t)}\,\frac{\cos(\bm v,\bm u_k)-\cos(\bm v,\bm\theta_k)\cos(\bm\theta_k,\bm u_k)}{\sqrt{1-\cos^2(\bm\theta_k,\bm u_k)}}.
    \label{eq:wn_schedule_exact_a_decomp}
\end{equation}

\textbf{First-order optimality approximation and measurement of \(a(t)\).}
Let \(\Loss_k(\bm\theta)\) denote the minibatch loss at step \(k\), with gradient \(\bm g_k=\nabla\Loss_k(\bm\theta_k)\).
A first-order expansion of \(\Loss_k(R\bm v)\) around \(\bm\theta_k=R\hat{\bm\theta}_k\) on the sphere gives
\[
    \Loss_k(R\bm v)-\Loss_k(\bm\theta_k)
    =R\bm g_k^\top(\bm v-\hat{\bm\theta}_k)+O\!\left(R^2\norm{\bm v-\hat{\bm\theta}_k}^2\right).
\]
Since the optimal direction \(\bm v\) lowers the objective relative to \(\hat{\bm\theta}_k\), to first order the tangential component \(\bm v-(\bm v^\top\hat{\bm\theta}_k)\hat{\bm\theta}_k\) aligns with the steepest-descent tangent \(-\bm g_k^\perp\):
\[
    \frac{\bm v-(\bm v^\top\hat{\bm\theta}_k)\hat{\bm\theta}_k}{\sqrt{1-(\bm v^\top\hat{\bm\theta}_k)^2}}
    =-\frac{\bm g_k^\perp}{\norm{\bm g_k^\perp}}+\bm\delta_k,
\]
where \(\bm\delta_k\) captures higher-order curvature.
Since \(\hat{\bm\theta}_k^\top\bm u_k^\perp=0\), we have the exact identity \(\bm v^\top\bm u_k^\perp=\bigl[\bm v-(\bm v^\top\hat{\bm\theta}_k)\hat{\bm\theta}_k\bigr]^\top\bm u_k^\perp\).
Substituting the tangential relation above, dropping \(\bm\delta_k\), and using \(\sqrt{1-(\bm v^\top\hat{\bm\theta}_k)^2}\approx 1\) in the unsaturated regime \((\bm v^\top\hat{\bm\theta}_k)^2\ll 1\) replaces \(\bm v\) by \(-\bm g_k^\perp/\norm{\bm g_k^\perp}\) and yields
\begin{equation}
    \frac{(-\bm g_k^\perp/\norm{\bm g_k^\perp})^\top\bm u_k^\perp}{R}
    =\sqrt{2b(t)}\,\frac{\cos(\bm u_k,-\bm g_k)-\cos(\bm\theta_k,\bm u_k)\cos(\bm\theta_k,-\bm g_k)}
    {\sqrt{1-\cos^2(\bm\theta_k,\bm u_k)}\,\sqrt{1-\cos^2(\bm\theta_k,-\bm g_k)}}.
    \label{eq:wn_schedule_exact_grad_a}
\end{equation}
In high dimensions, parameters are nearly orthogonal to updates and gradients: empirically, \(|\cos(\bm\theta_k,-\bm g_k)|\sim 10^{-4}\) and \(|\cos(\bm\theta_k,\bm u_k)|\ll 1\), making all radial cosine products and squares in Eq.~\eqref{eq:wn_schedule_exact_grad_a} negligible (within \(0.01\%\) of the simplified form below).
Dropping these higher-order radial terms reduces Eq.~\eqref{eq:wn_schedule_exact_grad_a} directly to our measured drift trajectory:
\begin{equation}
    a(t)=\sqrt{2b(t)}\,\cos(\bm u_k,-\bm g_k).
    \label{eq:wn_schedule_measured_a}
\end{equation}
Any constant or slowly varying scale factor in \(a(t)\) cancels in both \(\rho_a(t)=a(t)/\bar a_{\mathrm{rms}}(t)\) and \(\eta^\star(t)/\eta^\star(0)\).
Numerical quadrature of \((a(t),b(t))\) over \([0,t]\) then gives \(\bar a_{\mathrm{rms}}(t)\), \(\rho_a(t)\) in Eq.~\eqref{eq:wn_schedule_rms_def}, and \(\tau(t)=\int_0^t \frac{a(s)^2}{2b(s)}\,\dd s\), which are substituted into Eq.~\eqref{eq:wn_schedule_general_opt} with \(m_0^2=\alpha a(0)^2/[2b(0)]\) (\(\alpha=1{,}500\) steps) in Figure~\ref{fig:sec3_self_normalization}(c).

\subsection{Affine endpoint construction and rate-preservation properties}
\label{app:wn_schedule_affine_endpoints}
\label{app:wn_schedule_affine}

Let \(K\geq1\) be the integer decay horizon from Section~\ref{sec:wn_schedule_hart}.
For \(\alpha>0\), the inverse-square-root curve in Eq.~\eqref{eq:wn_schedule_shape} satisfies \(q(0)=1>q(K)>0\).
Writing an affine map as \(s q(k)+d\), prescribed endpoints \(\eta(0)=p\) and \(\eta(K)=\eta_{\min}\) require \(s+d=p\) and \(s q(K)+d=\eta_{\min}\), which uniquely determine
\[
    s=\frac{p-\eta_{\min}}{1-q(K)}>0,\qquad
    d=\eta_{\min}-s q(K),
\]
giving Eq.~\eqref{eq:wn_schedule_hart}.

\textbf{Uniqueness of Eq.~\eqref{eq:wn_schedule_affine_decrements} and minimum relative decay-rate distortion.}
For the continuous interpolation \(\eta_{\OurMethodMath}(t)=s q(t)+d\) on \(t\in[0,K]\), differentiating gives \(\dot\eta_{\OurMethodMath}(t)=s\dot q(t)<0\) and \(\ddot\eta_{\OurMethodMath}(t)=s\ddot q(t)>0\).
Dividing \(\dot\eta_{\OurMethodMath}(t_1)\) by \(\dot\eta_{\OurMethodMath}(t_2)\) cancels \(s>0\) and yields the first identity in Eq.~\eqref{eq:wn_schedule_affine_decrements}, while subtracting \(\eta_{\OurMethodMath}(j)\) from \(\eta_{\OurMethodMath}(i)\) and dividing by \(p-\eta_{\min}=s[1-q(K)]\) yields the second identity.
Conversely, if a smooth schedule \(\eta\in C^1([0,K])\) with \(\eta(0)=p\) and \(\eta(K)=\eta_{\min}\) satisfies \(\dot\eta(t_1)/\dot\eta(t_2)=\dot q(t_1)/\dot q(t_2)\) for all \(t_1,t_2\in[0,K]\), setting \(t_2=0\) gives \(\dot\eta(t)=[\dot\eta(0)/\dot q(0)]\dot q(t)\), which integrates under the two endpoint conditions uniquely to \(\eta_{\OurMethodMath}\).
Moreover, relative to the unconstrained profile \(\eta^\star(t):=p\,q(t)\) (whose total reduction on \([0,K]\) is \(\int_0^K\lvert\dot\eta^\star(t)\rvert\,\dd t=p[1-q(K)]\)), any \(\eta\in C^1([0,K])\) with \(\eta(0)=p\) and \(\eta(K)=\eta_{\min}\) satisfies \(\int_0^K(-\dot\eta(t))\,\dd t=p-\eta_{\min}\) and, by the Cauchy--Schwarz inequality,
\[
    \begin{aligned}
        \int_0^K\left(\frac{\dot\eta(t)-\dot\eta^\star(t)}{\dot\eta^\star(t)}\right)^{\!2}\lvert\dot\eta^\star(t)\rvert\,\dd t
        &=\int_0^K\frac{\dot\eta(t)^2}{\lvert\dot\eta^\star(t)\rvert}\,\dd t-2(p-\eta_{\min})+p\bigl[1-q(K)\bigr]\\
        &\geq\frac{\bigl[(p-\eta_{\min})-p(1-q(K))\bigr]^2}{p[1-q(K)]},
    \end{aligned}
\]
with equality if and only if \(\dot\eta(t)/\dot\eta^\star(t)\equiv\mathrm{const}\) on \([0,K]\), i.e., \(\eta=\eta_{\OurMethodMath}\).
Finally, \(\ddot\eta_{\OurMethodMath}(t)>0\) preserves strict convexity on \([0,K]\), Taylor expansion around \(t=K\) gives smooth first-order linear landing \(\eta_{\OurMethodMath}(t)-\eta_{\min}=-\dot\eta_{\OurMethodMath}(K)(K-t)+O\bigl((K-t)^2\bigr)\) without any derivative kink, and \(q(K)\to 0\) recovers \(p\,q(t)\) as \(K/\alpha\to\infty\) and \(\eta_{\min}\to 0\).

\subsection{Matching the discrete LR sum}
\label{app:wn_schedule_lr_sum}

Fix the warmup rule, decay horizon \(K\), time scale \(\alpha\), terminal LR \(\eta_{\min}\), and the executed decay indices \(\mathcal I\subseteq\{0,\ldots,K\}\) with \(n_{\mathrm d}=|\mathcal I|\).
Write the warmup values as \(c_i p+d_i\), where \(c_i,d_i\) are fixed independently of the peak \(p\).
For a target sum \(S_{\mathrm d}\) over all executed LR values, define
\[
    C_W:=\sum_{i\in\mathrm{warmup}}c_i,\qquad
    B_W:=\sum_{i\in\mathrm{warmup}}d_i,\qquad
    F_{\mathcal I}:=\sum_{k\in\mathcal I}f(k).
\]
Summing the warmup and Eq.~\eqref{eq:wn_schedule_hart} gives
\[
    S_{\mathrm d}=B_W+p(C_W+F_{\mathcal I})+\eta_{\min}(n_{\mathrm d}-F_{\mathcal I}),
\]
so whenever \(C_W+F_{\mathcal I}>0\) the peak has the unique solution
\begin{equation}
    p=\frac{S_{\mathrm d}-B_W
    -\eta_{\min}(n_{\mathrm d}-F_{\mathcal I})}{C_W+F_{\mathcal I}},
    \label{eq:wn_schedule_peak_calibration}
\end{equation}
subject to \(p>\eta_{\min}\) and prescribed warmup constraints.
For \(W\geq1\) linear warmup values \(\eta_{\mathrm{init}}+(p-\eta_{\mathrm{init}})i/W\) (\(i=0,\ldots,W-1\)) followed by all decay values \(k=0,\ldots,K\), we have \(C_W=(W-1)/2\), \(B_W=\eta_{\mathrm{init}}(W-C_W)\), and \(n_{\mathrm d}=K+1\).

\subsection{Uniqueness under convex last-iterate tail-sum zeroing}
\label{app:wn_schedule_tailsum}

Let \(r\in C^1([0,K])\) be a positive, strictly decreasing base profile normalized to \(r(0)=1\), and consider the continuous counterpart \(r(t)\int_t^K r(u)\,\dd u\) of the last-iterate bound-minimizing tail-sum schedule~\citep{defazio2024linear}.
Requiring this tail-sum profile to coincide identically on \([0,K]\) with an endpoint-affine transformation \(c\bigl(r(t)-r(K)\bigr)\) for some constant \(c>0\) is equivalent, after dividing by \(r(t)>0\), to
\[
    \int_t^K r(u)\,\dd u
    =c\left(1-\frac{r(K)}{r(t)}\right).
\]
Differentiating with respect to \(t\) gives \(-r(t)=c\,r(K)\dot r(t)/r(t)^2\), or equivalently
\[
    \frac{\dd}{\dd t}\!\left(\frac{1}{r(t)^2}\right)
    =-\frac{2\dot r(t)}{r(t)^3}
    =\frac{2}{c\,r(K)}=:\frac{1}{\alpha}>0.
\]
With \(r(0)=1\), integration uniquely yields \(r(t)=(1+t/\alpha)^{-1/2}=q(t)\) (with \(c=2\alpha/q(K)\)), for which
\[
    \frac{q(t)\int_t^K q(u)\,\dd u}{q(0)\int_0^K q(u)\,\dd u}
    =\frac{q(t)-q(K)}{1-q(K)}
    =f(t).
\]

\section{Experimental Settings and Evaluation Protocol}
\label{app:new_protocol}
\label{app:experimental_protocol}

This appendix details the experimental setup in Sections~\ref{sec:new_evaluation} and~\ref{sec:new_late_stage}.

\subsection{Training settings and comparison controls}
\label{app:new_protocol_controls}
Table~\ref{tab:train_setup} summarizes the training settings for the smaller-scale scheduler experiments.
The larger-scale comparison uses 790\,M activated and 35.7\,B total parameters, 500\,B training tokens, and 168 hours per run on 32 GPUs.
Within each fixed-norm scheduler comparison, the training setting is held fixed and the LR schedule is varied.

\begin{table}[htbp]
    \centering
    \caption{Training settings for the smaller-scale scheduler comparisons.
    The peak LRs specify the main comparison between cosine and \OurMethod; other schedule peaks are calibrated to the same cumulative LR.}
    \label{tab:train_setup}
    \begin{center}
        \begin{tabular}{lr}
            \toprule
            Configuration & Value \\
            \midrule
            Activated parameters & 110\,M \\
            Total parameters & 4.7\,B \\
            Training tokens & 128\,B \\
            Optimizer & Muon \\
            Global batch size & 128 \\
            Normalization & Frobenius norm \\
            Normalization epsilon & \(10^{-8}\) \\
            Scheduled optimizer updates & 32,047 \\
            Warmup updates & 3,040 \\
            Minimum LR & \(10^{-5}\) \\
            Cosine peak LR & \(1.07\times10^{-3}\) \\
            \OurMethod peak LR & \(1.95\times10^{-3}\) \\
            \OurMethod decay time scale & \(\alpha=K/8\) \\
            GPUs per run & 16 \\
            Wall-clock per complete run & 48 hours \\
            \bottomrule
        \end{tabular}
    \end{center}
\end{table}

\paragraph{Baseline peak-LR search.}
To ensure a well-tuned WN cosine baseline on the 110\,M/4.7\,B MoE model, we sweep six peak LRs over \([3.485\times 10^{-4},\, 1.9714\times 10^{-3}]\) while holding the remaining training settings fixed.
Figure~\ref{fig:cosine_baseline_lr_sweep} reports the terminal training and evaluation losses (defined in Appendix~\ref{app:new_protocol_loss}) together with strictly convex cubic fits in \(\log(\text{peak LR})\).
Based on the fitted evaluation-loss curve, we select the optimal peak LR of \(1.07\times 10^{-3}\) for the WN cosine baseline, from which the matched cumulative-LR peaks of the other decay schedules are calibrated.

\begin{figure}[htbp]
    \centering
    \includegraphics[width=\linewidth]{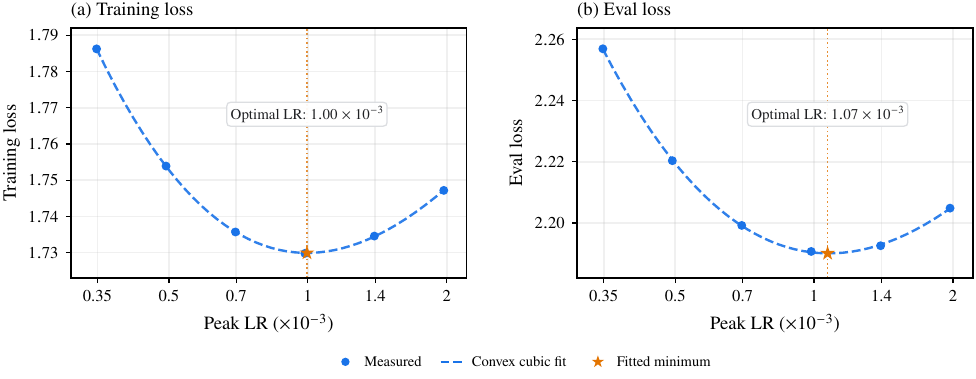}
    \caption{\textbf{Peak-LR search for the WN cosine baseline on the 110\,M/4.7\,B MoE model.}
    Terminal losses follow Appendix~\ref{app:new_protocol_loss} across six swept peak LRs; based on the evaluation-loss fit, we select \(1.07\times 10^{-3}\) as the baseline peak LR.}
    \label{fig:cosine_baseline_lr_sweep}
\end{figure}

While Figure~\ref{fig:hart_warmup_lr_refloss}(a) illustrates a linear warmup profile, all comparison controls apply to the complete executed optimizer-step schedule, including warmup, batch-size ramp, and the final logged update (step 32,046).
Within each fixed-norm comparison, the model, token budget, optimizer settings, batch-size schedule, constrained parameter blocks, and their radii are held fixed.
For matched-budget comparisons with cosine, peak calibration follows Appendix~\ref{app:wn_schedule_lr_sum}, holding the full-schedule LR sum, horizon, warmup duration, and terminal LR fixed while calibrating each family's peak LR.
In stage-wise interventions, a shared late schedule sets identical LR values at every shared optimizer step and recalibrates the prefix to match the full-schedule LR sum, whereas single-stage swaps splice the donor interval directly while leaving the recipient schedule unchanged elsewhere.
Table~\ref{tab:comparison_controls} summarizes the design controls across all schedule comparisons and temporal interventions.

\begin{table}[htbp]
    \centering
    \caption{Summary of design controls across schedule comparisons. The full-schedule LR sum includes warmup.}
    \label{tab:comparison_controls}
    \small
    \begin{tabular}{llll}
        \toprule
        Comparison & Peak LR & Terminal LR & Full LR sum \\
        \midrule
        Matched-budget families (Section~\ref{sec:new_evaluation_performance}) & Calibrated & Matched & Matched \\
        \OurMethod vs.\ Cycloid($\pi$), full runs (Section~\ref{sec:new_late_attribution}) & Matched & Matched & Matched \\
        Stage swaps (Section~\ref{sec:new_late_attribution}) & Matched & Matched & Unconstrained \\
        Shared late schedules (Section~\ref{sec:new_late_shared}) & Calibrated & Matched & Matched \\
        Late decay exponents (Section~\ref{sec:new_late_exponents}) & Matched & Matched & Within \(<1\%\) \\
        \bottomrule
    \end{tabular}
\end{table}

\subsection{Loss summaries}
\label{app:new_protocol_loss}

Tables~\ref{tab:multi_scheduler_compare} and~\ref{tab:fixrear_own_vs_shared_tail}, and the robustness comparison, use the same terminal loss convention.
For training loss, take the last 100 finite logged values in step order, denoted \(x_1,\ldots,x_{100}\), initialize \(z_1=x_1\), and recursively compute
\begin{equation}
    z_j=0.99z_{j-1}+0.01x_j,\qquad j=2,\ldots,100.
    \label{eq:new_protocol_loss}
\end{equation}
The reported value is \(z_{100}\), and evaluation loss is the last finite logged evaluation value without smoothing.
The smaller-scale comparison ends with training loss at step 32,046 and evaluation loss at step 32,000; the corresponding larger-scale steps are 62,602 and 62,000.

Training restarts are stitched by optimizer step, with the later-started segment taking precedence at duplicate steps.
For reference loss curves, we pair raw losses at common logged optimizer steps, subtract the specified reference, and apply a debiased EMA with decay 0.99 across the paired sequence:
\[
    \frac{\sum_{i=1}^{j}0.99^{j-i}d_i}{\sum_{i=1}^{j}0.99^{j-i}},
\]
where \(d_1,\ldots,d_j\) denote the paired differences.
The terminal reference loss points in Figure~\ref{fig:tail_landing_refloss} use the final value of this curve.
Figure~\ref{fig:fmr_stage_summary}(c) reports the terminal training-loss summary in Eq.~\eqref{eq:new_protocol_loss} for each of the eight runs, with stage-swap gaps computed as differences between these terminal values.

\subsection{Compared decay shapes}
\label{app:new_protocol_shapes}

Let \(k=0,\ldots,K\) index the decay stage as in
Eq.~\eqref{eq:wn_schedule_shape}, and write \(\tau=k/K\).
Thus \(\tau\in[0,1]\) is the fraction of the decay stage completed.
Table~\ref{tab:comparison_decay_shapes} gives the raw shape \(q(k)\)
for each schedule. All shapes satisfy \(q(0)=1\) and use the same
affine endpoint transformation and LR interpolation in
Eqs.~\eqref{eq:wn_schedule_shape}--\eqref{eq:wn_schedule_hart}, with their
own \(q\) substituted for the inverse-square-root curve.
The peak is then calibrated to match the cosine baseline's cumulative
LR, including warmup, following Appendix~\ref{app:wn_schedule_lr_sum}.
This holds the warmup duration and terminal LR fixed while calibrating each family's peak LR; the raw inverse-square-root curve in Figure~\ref{fig:hart_warmup_lr_refloss}(a) serves as an uncalibrated reference sharing \OurMethod's peak LR.

\begin{table}[htbp]
    \caption{Raw decay shapes before the common affine transformation, and
    the shape parameter selected for the smaller-scale rows of
    Table~\ref{tab:multi_scheduler_compare}. The time scale \(\alpha>0\), decay exponent
    \(\beta>0\), which is fixed for the two power families and rises to
    \(\beta_{\mathrm{end}}\) for variable power, the terminal angle
    \(0<\theta_1\leq\pi\), and the exponential decay parameter
    \(\rho>0\) control the respective curve shapes. A dash marks a shape
    with no free parameter.}
    \label{tab:comparison_decay_shapes}
\begin{center}
    \small
    \begin{tabular}{lll}
        \toprule
        Shape & \(q(k)\), with \(\tau=k/K\) & Selected \\
        \midrule
        Cosine & \(\tfrac12[1+\cos(\pi\tau)]\) & -- \\
        Linear & \(1-\tau\) & -- \\
        \addlinespace
        Half-cosine & \(\cos(\pi\tau/2)\) & -- \\
        Rear-cosine & \(1-\sin(\pi\tau/2)\) & -- \\
        Cycloid & \(\cos\theta,\quad
            \theta-\sin\theta=\tau(\theta_1-\sin\theta_1)\)
            & \(\theta_1=3\pi/4\) \\
        Double-cosine & \(q_{\mathrm{dc}}(\tau)\), defined below & -- \\
        \addlinespace
        Power of \(1-\tau\) & \((1-\tau)^{\beta}\) & \(\beta=2\) \\
        Power of \(\tau\) & \(1-\tau^{\beta}\) & \(\beta=1/5\) \\
        \addlinespace
        Logarithmic & \(1-\log(1+k/\alpha)\) & \(\alpha=K/32\) \\
        Exponential & \(\exp(-\rho\tau)\) & \(\rho=3.594\) \\
        Inverse square & \((1+k/\alpha)^{-2}\) & \(\alpha=K/2\) \\
        Inverse time & \((1+k/\alpha)^{-1}\) & \(\alpha=0.243K\) \\
        Variable power & \((1+k/\alpha)^{-[1/2+(\beta_{\mathrm{end}}-1/2)\tau]}\)
            & \(\alpha=K/8\), \(\beta_{\mathrm{end}}=1.771\) \\
        Inverse square root (\OurMethod) & \((1+k/\alpha)^{-1/2}\)
            & \(\alpha=K/8\) \\
        \bottomrule
    \end{tabular}
\end{center}
\end{table}

The square-root shape used outside
Table~\ref{tab:multi_scheduler_compare} is \(1-\tau^\beta\) with \(\beta=1/2\), and the
quadratic shape is \((1-\tau)^\beta\) with \(\beta=2\). In
Table~\ref{tab:multi_scheduler_compare} each schedule is named by the
shape in this table, written up to the affine transformation: the time
variable \(t\) absorbs an affine rescaling, so \(t=\tau\) for the
polynomial shapes and \(t=1+k/\alpha\) for the inverse-power and logarithmic ones, and \(t=\rho\tau\) for the exponential shape, and the endpoint normalization is left implicit.
Cosine, Half-cosine, Rear-cosine, and Cycloid retain their full names in Table~\ref{tab:multi_scheduler_compare}.

For the symmetric double-cosine schedule, the two cosine segments
join at \(\tau=1/2\) and \(q_{\mathrm{dc}}=1/2\):
\[
    q_{\mathrm{dc}}(\tau)=
    \begin{cases}
        \tfrac34+\tfrac14\cos(2\pi\tau), & 0\leq\tau\leq\tfrac12,\\
        \tfrac14+\tfrac14\cos(2\pi\tau-\pi), & \tfrac12<\tau\leq1.
    \end{cases}
\]
For Cycloid, \(\theta\in[0,\theta_1]\) is the unique
solution of the relation in the table. For variable power, the
exponent varies linearly from \(1/2\) to \(\beta_{\mathrm{end}}\).

\subsection{Downstream benchmarks}
\label{app:protocol_downstream_benchmarks}

Table~\ref{tab:multi_scheduler_compare} reports performance on
MMLU~\citep{hendrycks2021mmlu}, MMLU-Pro~\citep{wang2024mmlupro},
AGIEval~\citep{zhong2024agieval}, WinoGrande~\citep{sakaguchi2020winogrande},
BBH~\citep{suzgun2023bbh}, DROP~\citep{dua2019drop},
GSM8k~\citep{cobbe2021gsm8k}, MATH~\citep{hendrycks2021math},
HumanEval~\citep{chen2021humaneval}, MBPP+~\citep{liu2023evalplus},
and TriviaQA~\citep{joshi2017triviaqa}.

\subsection{Comparisons with WD cosine and WSD recipes}
\label{app:protocol_existing_recipes}

To complement Table~\ref{tab:multi_scheduler_compare}, Table~\ref{tab:rsi_existing_recipes} directly compares WN \OurMethod against the canonical WD cosine baseline and WD WSD on the 128\,B-token MoE budget (with WD coefficient \(\lambda=0.1\)).
All terminal training and evaluation losses follow the reporting convention in Appendix~\ref{app:new_protocol_loss}.

\begin{table}[htbp]
    \centering
    \caption{\textbf{Terminal losses across WN and WD schedules at 128\,B tokens.}
    Peak LRs are in units of \(10^{-3}\); terminal training and evaluation losses follow Appendix~\ref{app:new_protocol_loss}.}
    \label{tab:rsi_existing_recipes}
    \small
    \begin{adjustbox}{max width=\linewidth}
        \begin{tabular}{llrrr}
    \toprule
    Norm control & Schedule & Peak LR ($10^{-3}$) & Train ($\downarrow$) & Eval ($\downarrow$) \\
    \midrule
    WD & Cosine & 1.07 & 1.753 & 2.205 \\
    WD & WSD & 1.07 & 1.763 & 2.203 \\
    WN & \OurMethod & 1.95 & \textbf{1.730} & \textbf{2.174} \\
    \bottomrule
\end{tabular}

    \end{adjustbox}
\end{table}

\begin{figure}[htbp]
    \centering
    \includegraphics[width=\textwidth]{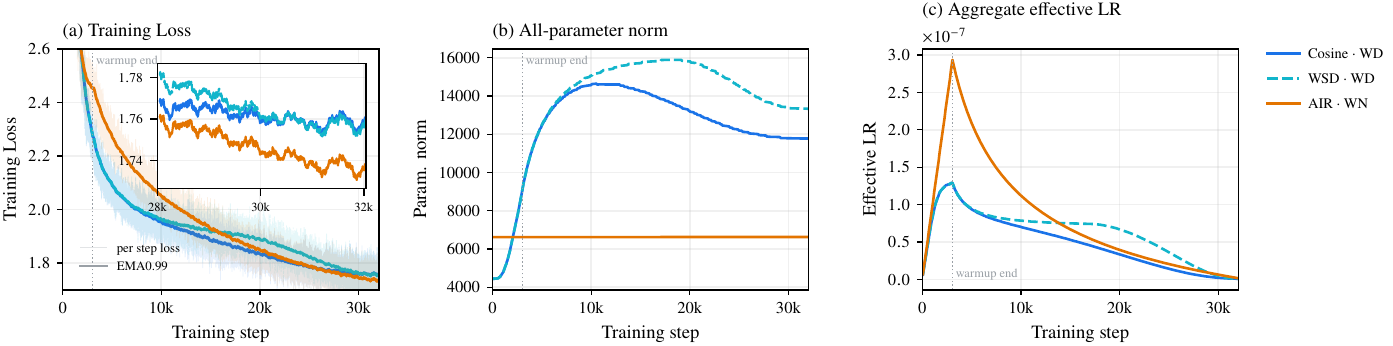}
    \caption{\textbf{WN \OurMethod achieves lower loss than standard WD cosine and WSD schedules through direct ELR control.}
    Panel~\textbf{(a)} plots EMA-0.99 training loss (inset: final 4k steps); panels~\textbf{(b,c)} show the all-parameter norm and aggregate ELR in Eq.~\eqref{eq:rsi_aggregate_elr}. Dotted vertical lines mark warmup end.}
    \label{fig:rsi_recipe_comparison}
\end{figure}

Figure~\ref{fig:rsi_recipe_comparison} compares the training-loss, all-parameter norm, and aggregate ELR trajectories from the same three trials as Table~\ref{tab:rsi_existing_recipes}.
Across all logged parameter blocks \(\mathcal B\), we compute the all-parameter root-sum-square norm and aggregate ELR as
\begin{equation}
    r_{\mathrm{agg}}(k)=\sqrt{\sum_{j\in\mathcal B}\norm{\bm\theta_{j,k}}^2},
    \qquad
    \eta_{\mathrm{agg}}^{\mathrm{eff}}(k)=\frac{\eta_k}{r_{\mathrm{agg}}(k)}.
    \label{eq:rsi_aggregate_elr}
\end{equation}
Unlike WD cosine (whose growing norm drives ELR down rapidly early on) and WD WSD (whose ELR plateaus before a late drop), WN \OurMethod decouples ELR from norm drift and sustains smooth convex annealing from a higher initial ELR to achieve the lowest terminal loss.

\subsubsection{Theoretical comparison of induced ELR trajectories}
\label{app:wd_wn_elr_theory}

\textbf{Equilibrium norm, square-root LR scaling, and relaxation lag under WD.}
To explain the parameter-norm and ELR trajectories in Figures~\ref{fig:elr_cosine_wsd}(b,c) and~\ref{fig:rsi_recipe_comparison}(b,c), consider decoupled WD in Eq.~\eqref{eq:principle_wd_update} under scale-invariant tangential updates satisfying \(\bm\theta_k^\top\bm u_k=0\), \(\norm{\bm u_k}=U>0\), and \(0\leq\lambda\eta_k<1\).
Taking squared Euclidean norms gives the exact discrete recurrence and, for small relative steps (\(\lambda\eta_k\ll 1\) and \(\eta_k U/r_k\ll 1\)), its continuous-time ODE:
\begin{equation}
    r_{k+1}^2=(1-\lambda\eta_k)^2 r_k^2+U^2\eta_k^2,
    \qquad
    \frac{\dd}{\dd t}\bigl(r(t)^2\bigr)
    =-2\lambda\eta(t)\,r(t)^2+U^2\eta(t)^2
    =2\lambda\eta(t)\bigl(r_\star(t)^2-r(t)^2\bigr).
    \label{eq:wd_norm_ode}
\end{equation}
Balancing tangential update growth against radial WD shrinkage defines an instantaneous equilibrium norm \(r_\star(t)\) scaling as \(\sqrt{\eta(t)/\lambda}\) (with squared norm \(r_\star(t)^2\propto\eta(t)/\lambda\)), so the corresponding equilibrium ELR scales with the square root of the nominal LR~\citep{li2020intrinsic,kosson2026wdtransfer,wen2026hyperball}:
\begin{equation}
    r_\star(t):=U\sqrt{\frac{\eta(t)}{2\lambda}}
    \propto\sqrt{\frac{\eta(t)}{\lambda}},
    \qquad
    \elr_\star(t):=\frac{\eta(t)}{r_\star(t)}
    =\frac{\sqrt{2\lambda\eta(t)}}{U}
    \propto\sqrt{\lambda\eta(t)}.
    \label{eq:wd_equilibrium_elr}
\end{equation}
Because \(r(t)^2\) relaxes toward \(r_\star(t)^2\) at rate \(2\lambda\eta(t)\) rather than adjusting instantaneously, integrating Eq.~\eqref{eq:wd_norm_ode} with cumulative nominal LR \(S(t):=\int_0^t\eta(s)\,\dd s\) yields delayed norm evolution:
\begin{equation}
    r(t)^2=r_0^2 e^{-2\lambda S(t)}
    +U^2\int_0^t e^{-2\lambda(S(t)-S(s))}\eta(s)^2\,\dd s,
    \qquad
    \elr_{\mathrm{WD}}(t)=\frac{\eta(t)}{r(t)}.
    \label{eq:wd_elr_integral}
\end{equation}
Comparing Eq.~\eqref{eq:wd_elr_integral} against fixed-norm WN (\(\elr_{\mathrm{WN}}(t)=\eta(t)/R\)) separates training into an early delayed-growth regime where WD and WN + \(1/\sqrt{t}\) behave similarly, and an equilibrated regime where they diverge.

\textbf{When trajectories are similar (\(2\lambda S(t)\ll 1\) and \(r_0\ll r_\star\)).}
Before cumulative WD equilibrates, such as early in the WSD stable phase (\(\eta(t)\equiv\eta_0\) with \(2\lambda\eta_0 t\ll 1\)) or near the cosine peak, \(1-e^{-2\lambda\eta_0 t}=2\lambda\eta_0 t+O(\lambda^2\eta_0^2 t^2)\) and \(r(t)^2\approx r_0^2+U^2\eta_0^2 t\) grows linearly with step \(t\).
Dividing \(\eta_0\) by \(r(t)\) shows that delayed norm growth under WD passively induces shifted \(1/\sqrt{t}\) ELR decay:
\begin{equation}
    \elr_{\mathrm{WD}}(t)
    \approx\frac{\eta_0/r_0}{\sqrt{1+U^2\eta_0^2 t/r_0^2}},
    \label{eq:wd_early_invsqrt}
\end{equation}
which matches WN under raw \(1/\sqrt{t}\) decay in Eq.~\eqref{eq:wn_schedule_optimal_lr} with \(\alpha=r_0^2/(U^2\eta_0^2)\).

\textbf{When trajectories differ (\(2\lambda S(t)\gtrsim 1\)).}
Once cumulative WD reaches the equilibration scale, norm saturation and equilibrium tracking cause WD schedules to depart from strictly convex \(1/\sqrt{t}\) decay, matching Figures~\ref{fig:elr_cosine_wsd}(b,c) and~\ref{fig:rsi_recipe_comparison}(b,c):
\begin{itemize}
    \item \emph{WD + WSD (drop, plateau, then late drop).}
    During the constant-LR stable phase (\(\eta(t)\equiv\eta_0\)), Eq.~\eqref{eq:wd_elr_integral} gives \(r(t)^2=r_0^2 e^{-2\lambda\eta_0 t}+r_\star^2(1-e^{-2\lambda\eta_0 t})\) with characteristic lag \(\tau_{\mathrm{eq}}=(2\lambda\eta_0)^{-1}\).
    Consequently, \(\elr_{\mathrm{WD+WSD}}(t)\) first drops along Eq.~\eqref{eq:wd_early_invsqrt} while \(r(t)\) grows (\(t\ll\tau_{\mathrm{eq}}\)), gradually flattens onto the positive equilibrium plateau \(\elr_\star=\sqrt{2\lambda\eta_0}/U\) as \(r(t)\to r_\star\) (\(t\gtrsim\tau_{\mathrm{eq}}\)), and drops again during WSD's late nominal decay stage, whereas WN + WSD holds ELR constant at \(\eta_0/R\) throughout the stable phase.

    \item \emph{WD + Cosine (early drop, then square-root-cosine decay).}
    Writing \(\eta_{\mathrm{cos}}(t)\approx p\cos^2\!\bigl(\frac{\pi t}{2K}\bigr)\) for \(\eta_{\min}\ll p\), initial norm growth drives a rapid early ELR drop, after which quasi-equilibrium tracking (\(|\dot\eta_{\mathrm{cos}}/\eta_{\mathrm{cos}}|\ll 2\lambda\eta_{\mathrm{cos}}(t)\)) gives \(r(t)\approx r_\star(t)=U\sqrt{\eta_{\mathrm{cos}}(t)/(2\lambda)}\).
    Substituting Eq.~\eqref{eq:wd_equilibrium_elr} cancels half the nominal decay power and yields a square-root-cosine trajectory:
    \begin{equation}
        \elr_{\mathrm{WD+Cos}}(t)
        \approx\frac{\sqrt{2\lambda}}{U}\sqrt{\eta_{\mathrm{cos}}(t)}
        \approx\frac{\sqrt{2\lambda p}}{U}\cos\!\left(\frac{\pi t}{2K}\right),
        \label{eq:wd_sqrt_cosine}
    \end{equation}
    which follows a concave profile on \((0,K)\) rather than strictly convex \(1/\sqrt{t}\) decay, before norm relaxation slows near \(\eta_{\min}\) and leaves a positive terminal norm \(r_{\mathrm{end}}>0\).

    \item \emph{WN + \OurMethod.}
    Because WD couples \(\eta(t)\) with delayed norm evolution \(r(t)\), it mimics \(1/\sqrt{t}\) ELR decay only transiently during early norm growth before plateauing under WSD or turning concave under Cosine.
    Fixing \(r_k\equiv R\) under WN removes this norm lag so \OurMethod directly sustains calibrated, strictly convex \(1/\sqrt{t}\) ELR annealing from a higher initial ELR to \(\eta_{\min}/R\) across the full horizon (Figure~\ref{fig:rsi_recipe_comparison}(c)).
\end{itemize}

\subsubsection{Analytical instantiation in isotropic noisy linear regression}
\label{app:wd_wn_linear_toy}

The norm and ELR dynamics in Eqs.~\eqref{eq:wd_norm_ode}--\eqref{eq:wd_sqrt_cosine} carry over directly to linear regression with isotropic Gaussian inputs and observation noise.
At each step \(k\), draw \(B\) independent samples \(\{(\bm x_{k,i},y_{k,i})\}_{i=1}^B\) from \(y=\bm x^\top\bm\theta^\star+\xi\) with \(\bm x\sim\mathcal N(\bm 0,\bm I_d)\) and independent noise \(\xi\sim\mathcal N(0,\sigma^2)\).
For a scale-invariant normalized predictor \(y=\bm x^\top(\bm\theta/\norm{\bm\theta})+\xi\) with normalized tangential updates (\(\norm{\bm u_k}=U\)), radial gradients vanish identically (\(\bm\theta_k^\top\nabla\Loss(\bm\theta_k)=0\)) and decoupled WD obeys Eqs.~\eqref{eq:wd_norm_ode}--\eqref{eq:wd_sqrt_cosine} verbatim.

For standard unnormalized half-squared loss \(\frac{1}{2B}\sum_{i=1}^B(\bm x_{k,i}^\top\bm\theta-y_{k,i})^2\) optimized by SGD with decoupled WD (\(\bm u_k=-\bm g_k\)), conditioning on \(\bm\theta_k\) over the step-\(k\) minibatch and noise gives the exact first and second gradient moments:
\begin{equation}
    \E[\bm g_k\mid\bm\theta_k]=\bm\theta_k-\bm\theta^\star,
    \qquad
    \E[\norm{\bm g_k}^2\mid\bm\theta_k]
    =\left(1+\frac{d+1}{B}\right)\norm{\bm\theta_k-\bm\theta^\star}^2
    +\frac{d\sigma^2}{B}.
    \label{eq:linear_toy_moments}
\end{equation}
Expanding \(\bm\theta_{k+1}=(1-\lambda\eta_k)\bm\theta_k-\eta_k\bm g_k\) around the isotropic optimum \(\bm\theta^\star=\bm 0\) collects the linear restoring term with WD into an exact conditional squared-norm recurrence:
\begin{equation}
    \E[r_{k+1}^2\mid\bm\theta_k]
    =\left[\bigl(1-(\lambda+1)\eta_k\bigr)^2+\frac{d+1}{B}\eta_k^2\right]r_k^2
    +\eta_k^2 U_{\mathrm{LR}}^2,
    \qquad
    U_{\mathrm{LR}}^2:=\frac{d\sigma^2}{B}.
    \label{eq:linear_toy_recurrence}
\end{equation}
In the small-step regime \(\frac{d+1}{B}\eta_k\ll 2(\lambda+1)\), taking full expectations yields the same linear ODE as Eq.~\eqref{eq:wd_norm_ode}, \(\frac{\dd}{\dd t}\E[r(t)^2]=-2\tilde\lambda\eta(t)\,\E[r(t)^2]+U_{\mathrm{LR}}^2\eta(t)^2\), where quadratic loss curvature shifts the contraction rate to \(\tilde\lambda:=\lambda+1\) and minibatch noise variance \(U_{\mathrm{LR}}^2=d\sigma^2/B\) acts as \(U^2\).
Substituting \((\tilde\lambda,U_{\mathrm{LR}})\) into Eqs.~\eqref{eq:wd_equilibrium_elr}--\eqref{eq:wd_sqrt_cosine} recovers the same equilibrium scaling \(r_\star(t)=U_{\mathrm{LR}}\sqrt{\eta(t)/(2\tilde\lambda)}\) and \(\elr_\star(t)=\sqrt{2\tilde\lambda\eta(t)}/U_{\mathrm{LR}}\), together with identical similarity and divergence regimes: passive shifted \(1/\sqrt{t}\) ELR decay matching WN + \(1/\sqrt{t}\) during early delayed norm growth (\(2\tilde\lambda S(t)\ll 1\) and \(r_0\ll r_\star\)), versus equilibration to a plateau under WSD or a concave \(\sqrt{\eta_{\mathrm{cos}}(t)}\) profile under Cosine once \(2\tilde\lambda S(t)\gtrsim 1\).

\subsection{Hyperparameter sweeps and LR sensitivity}
\label{app:new_protocol_downstream}

For the robustness sweep in Section~\ref{sec:new_evaluation_robustness}, we
report each family's LR sensitivity. For a family with configurations
\(i=1,\dots,n\) of terminal loss \(L_i\) and peak LR \(\eta_i\), the LR
sensitivity is
\begin{equation}
    S=\frac{1}{W_{\log}}\left(\frac{1}{n}\sum_{i=1}^{n}L_i-\min_{i}L_i\right),
    \qquad
    W_{\log}=\ln\frac{\max_i\eta_i}{\min_i\eta_i},
    \label{eq:lr_sensitivity}
\end{equation}
the average loss above the family's best configuration per e-fold of peak LR:
the numerator is the mean terminal loss above the family's best measured
configuration, and dividing by the log-span \(W_{\log}\) of the swept peak LR
normalizes for search width.
Table~\ref{tab:lr_family_robustness} reports \(S\) on training and evaluation
loss.

\begin{table}[htbp]
    \caption{\textbf{\OurMethod is the least sensitive to the peak LR on
    training loss.} Sensitivity is the mean loss above a family's best per e-fold of
    peak LR (Eq.~\eqref{eq:lr_sensitivity}).}
    \label{tab:lr_family_robustness}
\begin{center}
    \begin{adjustbox}{max width=\linewidth}
        {\setlength{\tabcolsep}{4pt}\renewcommand{\arraystretch}{1.2}%
\begin{tabular}{lcc|cc}
    \toprule
    Schedule & Configs & LR range & \multicolumn{2}{c}{LR sensitivity ($\downarrow$)} \\
    \cmidrule(lr){4-5}
    & & (e-folds) & Train & Eval \\
    \midrule
    Cycloid & 3 & 0.3461 & 0.0022 & \textbf{0.0017} \\
    $(1-t)^{\beta}$ & 3 & 0.4219 & 0.0160 & 0.0126 \\
    $1-t^{\beta}$ & 3 & 0.5630 & 0.0023 & 0.0039 \\
    $1-\log t$ & 3 & 0.2182 & 0.0047 & 0.0054 \\
    $e^{-t}$ & 4 & 0.7081 & 0.0113 & 0.0110 \\
    $1/t^{2}$ & 3 & 0.5630 & 0.0032 & 0.0021 \\
    $1/t$ & 5 & 0.5874 & 0.0019 & 0.0027 \\
    $t^{-\beta(t)}$ & 2 & 0.0427 & 0.0230 & 0.0234 \\
    \rowcolor{gray!15} $1/\sqrt{t}$ (\OurMethod) & 5 & 0.7506 & \textbf{0.0018} & 0.0018 \\
    \bottomrule
\end{tabular}%
}%

    \end{adjustbox}
\end{center}
\end{table}

Table~\ref{tab:lr_family_robustness} reports the configuration counts and logarithmic spans \(W_{\log}\), and Table~\ref{tab:protocol_sweep_grid} lists the swept shape settings and resulting peak-LR intervals.
Sensitivity \(S\) is computed directly from the measured configurations, while Figure~\ref{fig:lr_family_training_loss} overlays polynomial fits of degree \(\min(3,n-1)\) (constrained to be convex over the observed peak-LR interval when \(n\geq 3\)) for visualization.

\begin{table}[htbp]
    \caption{Measured shape settings and peak-LR ranges for the smaller-scale sweep.
    Decimal shape settings are rounded to three places; the other values are exact.
    Counts and logarithmic spans appear in Table~\ref{tab:lr_family_robustness}.}
    \label{tab:protocol_sweep_grid}
\begin{center}
    \small
    \begin{tabular}{lll}
        \toprule
        Family & Shape settings & Peak LR (\(10^{-3}\)) \\
        \midrule
        Cycloid & \(\theta_1\in\{\pi/2,3\pi/4,\pi\}\) & 1.377--1.946 \\
        \((1-t)^\beta\) & \(\beta\in\{2,3,4\}\) & 1.528--2.330 \\
        \(1-t^\beta\) & \(\beta\in\{1/2,1/3,1/5\}\) & 1.528--2.683 \\
        Logarithmic & \(\alpha/K\in\{1/8,1/16,1/32\}\) & 1.541--1.916 \\
        Exponential & \(\rho\in\{2,3.594,5.549,7\}\) & 1.489--3.023 \\
        Inverse square & \(\alpha/K\in\{1,1/2,1/4\}\) & 1.528--2.683 \\
        Inverse time & \(\alpha/K\in\{0.864,0.400,0.243,0.166,0.125\}\) & 1.382--2.487 \\
        Variable power & \(\beta_{\mathrm{end}}\in\{1.771,2\},\ \alpha=K/8\) & 1.946--2.031 \\
        \OurMethod & \(\alpha/K\in\{0.557,1/4,1/8,1/16,1/32\}\) & 1.382--2.928 \\
        \bottomrule
    \end{tabular}
\end{center}
\end{table}

\subsection{Stage swaps}
\label{app:protocol_stage_swaps}
Figure~\ref{fig:fmr_stage_summary} compares \OurMethod with Cycloid at \(\theta_1=\pi\), which shares its peak LR, warmup, terminal LR, and full-schedule cumulative LR.
This comparator differs from the evaluation-selected Cycloid at \(\theta_1=3\pi/4\) in Table~\ref{tab:multi_scheduler_compare}.
The two intersections occur at approximately global steps 4,346 and 16,219 and define the early, middle, and late intervals.
Each swap replaces the donor schedule on the selected interval while keeping the recipient schedule unchanged outside it without rescaling.
The terminal summary and reference loss smoothing follow Appendix~\ref{app:new_protocol_loss}.

\subsection{Late decay exponents}
\label{app:protocol_late_exponents}
The exponent interventions in Eq.~\eqref{eq:new_late_gamma} switch at global step 25,955, corresponding to \(K_s/K\simeq0.79\) of the decay stage after warmup.
They use \(\gamma\in\{0.5,0.75,1,2,3\}\), sharing the same earlier schedule, junction LR, and configured terminal LR.
Across all five exponents, trapezoidal integration of the recorded LR traces remains within 1\,\% of the \OurMethod reference.
Figure~\ref{fig:tail_landing_refloss} reports same-step loss differences to the full-run \OurMethod reference with the full-sequence EMA convention in Appendix~\ref{app:new_protocol_loss}.

\section{Bounds for Early- and Late-Schedule Differences}
\label{app:schedule_comparisons}

\subsection{Common setting}
\label{app:comparison_setting}

We use the continuous relaxation of the decay-step coordinate \(k\in[0,K]\), with a fixed split \(0\leq K_s<K\).
For \(i\in\{1,2\}\), the expected alignment \(m_i\) follows the dynamics from Section~\ref{sec:wn_schedule} with constant curvature \(b>0\) (Section~\ref{sec:wn_schedule_optimal}),
\[
    \dot m_i(k)=a(k)\eta_i(k)-b\eta_i(k)^2m_i(k),
    \qquad m_i(0)=m_0\geq0,
\]
common bounded, measurable trajectory \(a(k)>0\), and bounded, measurable, nonnegative schedules \(\eta_i\).
Dots denote derivatives with respect to the continuous step coordinate \(k\).
All expectations condition on the same initial history \(\hist_0\) from Eq.~\eqref{eq:wn_schedule_alignment} throughout \([0,K]\).
Write \(\Delta m=m_1-m_2\) and \(\Delta\eta=\eta_1-\eta_2\).
A shared early schedule and the common initial alignment give \(\Delta m(K_s)=0\) by uniqueness of the linear ODE.

For each schedule, the retention factor from Eq.~\eqref{eq:wn_schedule_integral} is
\begin{equation}
    \begin{aligned}
        H_i(K,k)&=\exp\!\left[-b\int_k^K\eta_i(s)^2\,\dd s\right],\\
        H_{12}(K,k)&=\sqrt{H_1(K,k)H_2(K,k)}.
    \end{aligned}
    \label{eq:new_pair_retention}
\end{equation}
The factor \(H_{12}\) retains differences under the average retraction rate \(\tfrac12 b\bigl(\eta_1(s)^2+\eta_2(s)^2\bigr)\).
When the late schedules coincide, \(H_1=H_2=H_{12}=H\) on \([K_s,K]\).

\subsection{Varying early-stage schedules}
\label{app:comparison_common_tail}

If \(\eta_1=\eta_2=\eta\) on \([K_s,K]\), subtracting the two state equations cancels \(a(k)\eta(k)\) and gives
\[
    \frac{\dd\Delta m(k)}{\dd k}
    =-b\eta(k)^2\Delta m(k).
\]
Multiplying by \(\exp\!\bigl[b\int_{K_s}^k\eta(s)^2\,\dd s\bigr]\) and integrating from \(K_s\) to \(K\) yields
\begin{equation}
    \Delta m(K)=H(K,K_s)\Delta m(K_s).
    \label{eq:comparison_common_tail_proof}
\end{equation}
Since \(0<H(K,K_s)\leq1\), Eq.~\eqref{eq:comparison_common_tail_proof} proves Eq.~\eqref{eq:new_common_tail} for any \(\Delta m(K_s)\in\R\), preserving the sign and expected-alignment ordering while contracting the gap whenever \(\eta\) is nonzero on \([K_s,K]\).

\paragraph{Discrete counterpart.}
For integers \(K_s,K\) and the one-step alignment update \(m_{i,k+1}=m_{i,k}+a_k\eta_{i,k}-b\eta_{i,k}^2 m_{i,k}\) with \(b\eta_{i,k}^2<1\), a shared tail \(\eta_{1,k}=\eta_{2,k}=\eta_k\) for \(k=K_s,\ldots,K-1\) gives \(\Delta m_{k+1}=(1-b\eta_k^2)\Delta m_k\).
Iteration yields
\[
    \Delta m_K=\Delta m_{K_s}
    \prod_{k=K_s}^{K-1}(1-b\eta_k^2),
\]
preserving the same sign-preserving attenuation property in discrete time.

\subsection{Varying late-stage schedules}
\label{app:comparison_different_tails}

On \([K_s,K]\), suppressing the argument \(k\), the algebraic identity
\[
    \eta_1^2m_1-\eta_2^2m_2
    =\frac{\eta_1^2+\eta_2^2}{2}\Delta m
    +\frac{(\eta_1+\eta_2)(m_1+m_2)}{2}\Delta\eta
\]
gives the exact difference equation
\begin{equation}
    \frac{\dd\Delta m}{\dd k}
    =-\frac{b}{2}(\eta_1^2+\eta_2^2)\Delta m
    +\left[a-\frac{b}{2}(\eta_1+\eta_2)(m_1+m_2)\right]\Delta\eta.
    \label{eq:comparison_difference_ode}
\end{equation}
Using the integrating factor associated with the first term yields
\begin{align}
    \Delta m(K)
    &=H_{12}(K,K_s)\Delta m(K_s)\nonumber\\
    &\quad+\int_{K_s}^K H_{12}(K,k)
    \left[a(k)-\frac{b}{2}\bigl(\eta_1(k)+\eta_2(k)\bigr)\bigl(m_1(k)+m_2(k)\bigr)\right]
    \Delta\eta(k)\,\dd k.
    \label{eq:comparison_full_identity}
\end{align}
The first term propagates the entering difference, and the second accumulates differences caused by the later LR choices.
Equation~\eqref{eq:comparison_full_identity} holds globally for any pair of admissible schedules under the alignment ODE.

Let \(P\geq\max_{i\in\{1,2\}}\sup_{[K_s,K]}\eta_i\), \(M\geq\max_{i\in\{1,2\}}\sup_{[K_s,K]}m_i\), and \(A:=\sup_{[K_s,K]}a<\infty\) denote common upper bounds on \([K_s,K]\).
The integral solution preserves nonnegative states, and \(\dot m_i(k)\leq AP\), so \(M=\max_i m_i(K_s)+AP(K-K_s)\) is a valid finite choice (with \(M=1\) also valid when alignments stay in \([0,1]\)).
Since \(0\leq\frac{b}{2}\bigl(\eta_1(k)+\eta_2(k)\bigr)\bigl(m_1(k)+m_2(k)\bigr)\leq 2bPM\), the bracketed factor in Eq.~\eqref{eq:comparison_full_identity} lies between \(a(k)-2bPM\) and \(a(k)\) at each \(k\in[K_s,K]\), and its magnitude is bounded by
\[
    L_{12}=\sup_{k\in[K_s,K]}\max\bigl\{a(k),\,\bigl|a(k)-2bPM\bigr|\bigr\}.
\]
This preserves the cancellation between alignment gain and retraction before taking absolute values.
When only the late-stage schedule varies, the shared early schedule gives \(\Delta m(K_s)=0\), and applying the triangle inequality to Eq.~\eqref{eq:comparison_full_identity} proves Eq.~\eqref{eq:new_tail_distance}:
\[
    |\Delta m(K)|\leq L_{12}\int_{K_s}^K
    H_{12}(K,k)|\Delta\eta(k)|\,\dd k.
\]

If both stages vary, retaining the first term in Eq.~\eqref{eq:comparison_full_identity} with \(H_{12}>0\) proves Eq.~\eqref{eq:new_full_schedule_bound} for arbitrary entering differences \(\Delta m(K_s)\):
\[
    |\Delta m(K)|\leq H_{12}(K,K_s)|\Delta m(K_s)|
    +L_{12}\int_{K_s}^K H_{12}(K,k)|\Delta\eta(k)|\,\dd k.
\]
Together, these comparisons characterize how LR schedules promote expected directional alignment under common \(a(k)\) and \(b\), providing a mechanism for improved training loss.

\subsection{Slope bounds and endpoint information}
\label{app:comparison_slopes}

Assume \(\Delta m(K_s)=0\), absolutely continuous late schedules, and a common terminal LR \(\eta_1(K)=\eta_2(K)\).
Let
\(\varepsilon_1=\operatorname*{ess\,sup}_{[K_s,K]}|\dot\eta_1-\dot\eta_2|<\infty\).
Subtracting the terminal values and integrating the derivative difference gives
\[
    \Delta\eta(k)=-\int_k^K
    (\dot\eta_1(s)-\dot\eta_2(s))\,\dd s,
    \qquad |\Delta\eta(k)|\leq\varepsilon_1(K-k).
\]
Substitution into Eq.~\eqref{eq:new_tail_distance}, followed by \(H_{12}\leq1\), yields
\begin{align}
    |\Delta m(K)|
    &\leq L_{12}\varepsilon_1
    \int_{K_s}^K H_{12}(K,k)(K-k)\,\dd k
    \label{eq:comparison_slope_proof}\\
    &\leq\frac{L_{12}\varepsilon_1(K-K_s)^2}{2}.
    \label{eq:new_tail_slope}
\end{align}
If the schedules also share their LR at \(K_s\), integration from the left endpoint gives
\(|\Delta\eta(k)|\leq\varepsilon_1\min\{k-K_s,K-k\}\).
The integral of this envelope is \(\varepsilon_1(K-K_s)^2/4\), hence
\[
    |\Delta m(K)|\leq\frac{L_{12}\varepsilon_1(K-K_s)^2}{4}.
\]
This sharper \(1/4\) factor applies whenever the schedules are continuous across the switch \(K_s\).

\paragraph{Applicability to the late-stage ablation.}
The schedules in Eq.~\eqref{eq:new_late_gamma} are bounded and absolutely continuous for every \(\gamma>0\), so the LR-difference bound in Eq.~\eqref{eq:new_tail_distance} applies under the stated alignment dynamics.
Differentiating Eq.~\eqref{eq:new_late_gamma} directly gives
\[
    \dot\eta_\gamma(k)
    =-\frac{\gamma[\eta_{\OurMethodMath}(K_s)-\eta_{\min}]}{K-K_s}
    \left(\frac{K-k}{K-K_s}\right)^{\gamma-1},\qquad k<K.
\]
Its magnitude diverges at \(K\) when \(0<\gamma<1\), is constant when \(\gamma=1\), and vanishes there when \(\gamma>1\).
Consequently, comparisons involving \(0<\gamma<1\) are governed directly by the general \(L^1\) bound in Eq.~\eqref{eq:new_tail_distance}, whereas the slope bound in Eq.~\eqref{eq:new_tail_slope} applies to smooth tails with \(\gamma\geq 1\).
This bound connects the exponent intervention to expected directional alignment, complementing the empirical loss comparison in Section~\ref{sec:new_late_exponents}.

\paragraph{Knowing only the terminal slope.}
For twice continuously differentiable schedules with a common terminal LR and \(\Delta m(K_s)=0\), define
\(\delta_K=|\dot\eta_1(K)-\dot\eta_2(K)|\) and
\(C_2=\sup_{[K_s,K]}|\ddot\eta_1-\ddot\eta_2|\).
These measure the terminal slope difference and the curvature difference over the full interval.
Taylor expansion with integral remainder gives
\[
    |\Delta\eta(k)|\leq\delta_K(K-k)+\frac{C_2}{2}(K-k)^2.
\]
Substitution into the LR-difference bound yields
\[
    |\Delta m(K)|\leq L_{12}
    \left(\frac{\delta_K(K-K_s)^2}{2}+\frac{C_2(K-K_s)^3}{6}\right).
\]
Equal terminal slopes remove the first contribution, while the curvature term captures differences across the interior of the schedules.

\subsection{Affine power curves and the scope of the bounds}
\label{app:comparison_power_curves}

As an auxiliary example on \([K_s,K]\), fix a common initial LR \(P>\eta_{\min}\geq0\) and a time scale \(\alpha_{\mathrm{tail}}>0\).
For exponent \(\beta>0\), define
\[
    q_\beta(k)=\left(1+\frac{k-K_s}{\alpha_{\mathrm{tail}}}\right)^{-\beta},
    \qquad
    \eta_\beta(k)=\eta_{\min}+(P-\eta_{\min})
    \frac{q_\beta(k)-q_\beta(K)}{1-q_\beta(K)}.
\]
This affine inverse-power family shares its initial and terminal LRs, with \(\beta=1/2\) giving an inverse-square-root shape; whereas Eq.~\eqref{eq:new_late_gamma} varies the terminal landing power \((K-k)^\gamma\), \(\eta_\beta\) applies an affine inverse-power decay across \([K_s,K]\).
Direct differentiation gives
\[
    |\dot\eta_\beta(k)|=
    \frac{P-\eta_{\min}}{\alpha_{\mathrm{tail}}}
    \frac{\beta\bigl(1+(k-K_s)/\alpha_{\mathrm{tail}}\bigr)^{-\beta-1}}
    {1-\bigl(1+(K-K_s)/\alpha_{\mathrm{tail}}\bigr)^{-\beta}}.
\]
The terminal slope follows by setting \(k=K\).
On a compact interval of positive exponents, the slope is jointly continuous in \(\beta\) and \(k\), so close exponents give uniformly close slopes.
Combined with the preceding bounds, this continuity shows that nearby exponents produce nearby terminal expected alignments from a common entering state.
When the early stage also varies, Eq.~\eqref{eq:comparison_full_identity} incorporates the resulting entering-state difference.

\subsection{Infinitesimal sensitivity to the time of an LR change}
\label{app:comparison_sensitivity}

For an admissible family \(\eta_\varepsilon=\eta+\varepsilon\delta\eta\) with bounded perturbation \(\delta\eta\) and a common initial state, let \(\delta m=\partial_\varepsilon m_\varepsilon|_{\varepsilon=0}\) denote the first-order change in expected alignment.
Differentiating the integral solution with respect to \(\varepsilon\) at \(\varepsilon=0\) (justified by uniform boundedness of \(\eta\), \(\delta\eta\), and \(a\) on \([0,K]\)) gives an absolutely continuous \(\delta m\) satisfying
\[
    \frac{\dd\delta m(k)}{\dd k}
    =-b\eta(k)^2\delta m(k)+\bigl[a(k)-2b\eta(k)m(k)\bigr]\delta\eta(k),
    \qquad \delta m(0)=0.
\]
Applying the integrating factor yields
\[
    \delta m(K)=\int_0^K H(K,k)
    \bigl[a(k)-2b\eta(k)m(k)\bigr]\delta\eta(k)\,\dd k.
\]
Later perturbations are retained more strongly through \(H(K,k)\), while \(a(k)-2b\eta(k)m(k)\) determines their local alignment gain, identifying when an LR change most strongly influences terminal alignment.

\end{document}